\documentclass[sigconf, nonacm]{acmart}

\newcommand\vldbdoi{XX.XX/XXX.XX}
\newcommand\vldbpages{XXX-XXX}
\newcommand\vldbvolume{15}
\newcommand\vldbissue{11}
\newcommand\vldbyear{2022}
\newcommand\vldbauthors{Haoteng Yin, Muhan Zhang, Yanbang Wang, Jianguo Wang, Pan Li}
\newcommand\vldbtitle{\shorttitle} 
\newcommand\vldbavailabilityurl{https://github.com/Graph-COM/SUREL.git}
\newcommand\vldbpagestyle{plain} 

\usepackage{graphicx}
\usepackage{subfigure}
\usepackage{multirow}
\usepackage{amsmath,amsthm}
\usepackage[english]{babel}

\usepackage{xspace}
\newcommand{\proj}{SUREL\xspace}

\usepackage[ruled,linesnumbered]{algorithm2e}

\usepackage{enumitem}
\setlist[itemize]{align=parleft,left=0pt..1em}

\usepackage{color,xcolor,colortbl}
\definecolor{Gray}{gray}{0.9}
\newcommand{\ccolor}{\cellcolor{Gray}}

\renewcommand{\vec}[1]{\boldsymbol{\mathrm{#1}}}

\providecommand{\vh}{\ensuremath{\vec{h}}}
\newcommand{\RR}{\mathbb{R}}
\newcommand{\ZZ}{\mathbb{Z}}

\theoremstyle{definition}
\newtheorem{definition}{Definition}[section]

\begin{document}
\title{Algorithm and System Co-design for Efficient Subgraph-based Graph Representation Learning}

\author{Haoteng Yin$^{\dagger}$, Muhan Zhang$^{\ddagger}$, Yanbang Wang$^\S$, Jianguo Wang$^\dagger$, Pan Li$^{\dagger}$}
\affiliation{
{{$^\dagger$}{Department of Computer Science, Purdue University}}~~~~~$^\ddagger$ Institute for Artificial Intelligence, Peking University\\
 {{$^{\S}$}{Department of Computer Science, Cornell University}}}
\affiliation{
$^\dagger$\{yinht, csjgwang, panli\}@purdue.edu $^\ddagger$ muhan@pku.edu.cn $^\S$ ywangdr@cs.cornell.edu
}

\begin{abstract}
Subgraph-based graph representation learning (SGRL) has been recently proposed to deal with some fundamental challenges encountered by canonical graph neural networks (GNNs), and has demonstrated advantages in many important data science applications such as link, relation and motif prediction. However, current SGRL approaches suffer from scalability issues since they require extracting subgraphs for each training or test query. Recent solutions that scale up canonical GNNs may not apply to SGRL. Here, we propose a novel framework \proj for scalable SGRL by co-designing the learning algorithm and its system support. \proj adopts walk-based decomposition of subgraphs and reuses the walks to form subgraphs, which substantially reduces the redundancy of subgraph extraction and supports parallel computation. Experiments over six homogeneous, heterogeneous and higher-order graphs with millions of nodes and edges demonstrate the effectiveness and scalability of \proj. In particular, compared to SGRL baselines, \proj achieves 10$\times$ speed-up with comparable or even better prediction performance; while compared to canonical GNNs, \proj achieves 50\% prediction accuracy improvement.
\end{abstract}

\maketitle

\pagestyle{\vldbpagestyle}
\begingroup\small\noindent\raggedright\textbf{PVLDB Reference Format:}\\
\vldbauthors. \vldbtitle. PVLDB, \vldbvolume(\vldbissue): \vldbpages, \vldbyear.\\
\href{https://doi.org/\vldbdoi}{doi:\vldbdoi}
\endgroup
\begingroup
\renewcommand\thefootnote{}\footnote{\noindent
This work is licensed under the Creative Commons BY-NC-ND 4.0 International License. Visit \url{https://creativecommons.org/licenses/by-nc-nd/4.0/} to view a copy of this license. For any use beyond those covered by this license, obtain permission by emailing \href{mailto:info@vldb.org}{info@vldb.org}. Copyright is held by the owner/author(s). Publication rights licensed to the VLDB Endowment. \\
\raggedright Proceedings of the VLDB Endowment, Vol. \vldbvolume, No. \vldbissue\ %
ISSN 2150-8097. \\
\href{https://doi.org/\vldbdoi}{doi:\vldbdoi} \\
}\addtocounter{footnote}{-1}\endgroup

\ifdefempty{\vldbavailabilityurl}{}{
\vspace{.3cm}
\begingroup\small\noindent\raggedright\textbf{PVLDB Artifact Availability:}\\
The source code, data, and/or other artifacts have been made available at \url{\vldbavailabilityurl}.
\endgroup
}

\section{Introduction}\label{sec:intro}
Graph-structured data is prevalent to model relations and interactions between elements in real-world applications~\cite{koller2007introduction}. Graph representation learning (GRL) aims to learn representations of graph-structured data and has recently become a hot research topic~\cite{hamilton2020graph}. Previous works on GRL focus on either model design or system design while very few works jointly consider them. Works on model design tend to propose more expressive, generalizable and robust GRL models while paying less attention to their deployment~\cite{velivckovic2017graph,morris2019weisfeiler}. Hence, many theoretically powerful models can hardly apply to large real-world graphs. On the other hand, research on system design focuses on system-level techniques for better model development, such as graph partitioning~\cite{chiang2019cluster}, sub-sampling~\cite{hamilton2017inductive,zeng2019graphsaint} and pipelining~\cite{zhang2020agl,jia2020improving,thorpe2021dorylus,zhang2021grain}. However, they only consider basic GRL models, in particular graph neural network (GNN) models, yet often overlook their modeling limitations to solve practical GRL tasks.

\begin{figure}[t]
\centering
\includegraphics[width=0.38\textwidth]{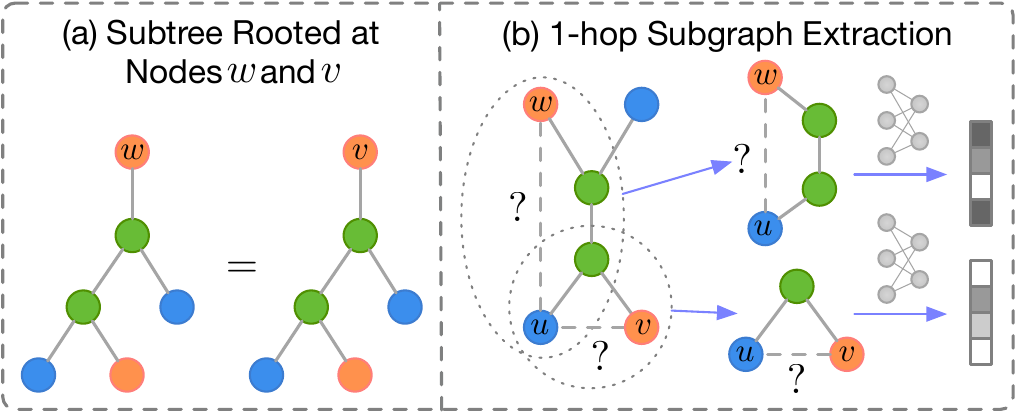}
\vspace{-3mm}
\caption{\small{A Toy Example of SGRL: the task is to predict whether $uw$ or $uv$ is more likely to form a link. Ideally, if this comes from a social network, $uv$ is more likely linked because they share a common neighbor. However, canonical GNNs cannot tell such difference since $w$ and $v$ share the same subtree structures resulting in the same representation~\citep{xu2019powerful}. SGRL solves this problem by extracting a subgraph patch around each queried node pair. Prediction based on the subgraph representation provides much better performance than canonical GNNs~\citep{zhang2018link,zhang2021labeling}.}}
\vspace{-4mm}
\label{fig:example}
\end{figure}

Canonical GNNs~\cite{kipf2016semi,hamilton2017inductive} share a common framework: each node is associated with a vector representation that gets iteratively updated by aggregating the representations from its neighboring nodes via graph convolution layers. The final prediction is made by combining the representations of nodes of interest. Although recent successes in system research have greatly pumped up the efficiency~\cite{fey2019fast,wang2019deep}, the GNN framework intrinsically suffers from three modeling limitations. First, information may be over-squashed into a single node representation that results in subpar performance when multiple tasks are associated, e.g. to predict multiple relations or links attached to the same node~\cite{epasto2019single,alon2021bottleneck}. Second, canonical GNNs cannot capture intra-node distance information due to limited expressive power~\cite{li2020distance,wang2022equivariant}, and thus fail to make predictions over a set of nodes (See Fig.~\ref{fig:example}a), such as substructure counting~\cite{bouritsas2020improving,zhengdao2020can} and higher-order pattern prediction~\cite{srinivasan2019equivalence,zhang2021labeling}. Third, the depth of GNNs is entangled with the range of the receptive field. For more non-linearity, using deeper GNNs comes with a larger but possibly unnecessary receptive field, which poses the risk of contaminating the representations with irrelevant information~\cite{huang2020graph,zeng2021decoupling}.

Recently, subgraph-based GRL (SGRL) has emerged as a new trend and has shown superior performance in tasks such as link prediction \cite{zhang2018link,zhang2021labeling}, relation prediction \cite{teru2020inductive}, higher-order pattern prediction \cite{li2020distance,liu2021neural}, temporal network modeling~\cite{wang2021inductive}, recommender systems~\cite{zhang2020inductive}, graph meta-learning \cite{huang2020graph}, and subgraph matching~\cite{liu2020neural,lou2020neural} and prediction~\cite{wang2021glass}. Different from canonical GNNs, SGRL extracts a subgraph patch for each training and test query and learns the representation of the extracted patch for final prediction (See Fig.~\ref{fig:example}b). For example, SEAL \cite{zhang2018link,zhang2021labeling} learns the representation of a subgraph around a given node pair to predict the link between them. This framework fundamentally overcomes the above three limitations. First, subgraph extraction allows decoupling the contributions made by a node to different queries, which prevents information over-squashing. Second, subgraph patches can be paired with distance-related features that favor prediction over a set of nodes~\cite{li2020distance,zhang2021labeling}. Third, subgraph extraction disentangles model depth and range of receptive field, which allows learning a rather non-linear model with only relevant local subgraphs as input.

Despite their importance, the SGRL framework has not received as much attention as the canonical GNN framework in the system research community. The underlying challenge comes from the subgraph extraction step in SGRL, which can be rather irregular and time-consuming. Specifically, SGRL requires to materialize a subgraph patch for each query during training and inference. Previous works of SGRL typically extract subgraphs offline for all such queries~\cite{zhang2018link,liu2021neural}, but it is not scalable for large graphs due to extensive memory need. Meanwhile, the online extraction~\cite{zeng2021decoupling} is not an option as it requires considerable processing time. The irregularity of subgraphs further makes it difficult to efficiently handle the extraction process in both cases.

Here, we aim to fill the gap by designing a novel computational framework \proj, to support SGRL over large graphs. \proj consists of a new system-friendly learning algorithm for SGRL and a scalable system to support this algorithm. The crucial design of \proj is to reduce the overhead caused by the online subgraph extraction, which all current SGRL approaches suffer from.

The key idea behind \proj is to break (and down-sample) subgraphs into random walks of regular size that can be easily sampled and, more importantly, reused among different queries. To compensate for the missing structural information after subgraph decomposition, we introduce relative position encoding (RPE), an intra-node distance feature that records the position of each node in the sampled subgraph. Specifically, for each node $u$ in the network, \proj collects a certain number of random walk starting from $u$. Each node appearing in these walks uses its landing counts at each step as the RPE vector. Overall, the set of collected walks paired with RPEs can be viewed as a subgraph patch centered at $u$. The complexity of the above process is linear with the number of nodes, and can be done in parallel and offline. For training and inference, given a queried node set $Q$, \proj first groups the sampled walks originated from all nodes in $Q$. Then, it implicitly joins the subgraph patches centered at each node in $Q$ by combining their node-level RPEs into a query-level RPE for each node associated in the grouped walks, which can also be executed in full parallel. Finally, \proj uses neural networks to learn the representation of the joined set of walks attached with query-level RPEs for final prediction. Since these walks are regular, the training process can be done quickly by GPU. The system architecture of \proj is illustrated in Fig.~\ref{fig:sgrl}.

Our contributions can be summarized as follows: (1) \textbf{A Novel System-Friendly Algorithm.} We propose the first scalable algorithm for SGRL tasks by adopting a novel walk-based computation framework. This framework uses regular data structures and allows extreme system acceleration. (2) \textbf{Dedicated System Support (\href{\vldbavailabilityurl}{Open-source}).} We design \proj to support the proposed algorithm. It can rapidly sample walks, encode positional features, and join them to represent multiple subgraphs in parallel. \proj adopts many system optimization techniques including parallelization, memory management, load balancing, etc. (3) \textbf{High Performance and Efficiency.} We evaluate \proj on link/relation/motif three prediction tasks over 6 real-world graphs of millions of nodes/edges. \proj significantly outperform the current SGRL approaches, and executes 10$\times$ faster in training and testing. Meanwhile, benefiting from the SGRL essence, \proj outperforms canonical GNNs by a great margin on prediction performance (almost $50\%$ in all tasks).
\section{Preliminaries and Related Works}
In this section, we set up notations, formulate the SGRL problem and review some related works.

\subsection{Notations}
\begin{definition}[Graph-structured data] Let $\mathcal{G}=(\mathcal{V},\mathcal{E},X)$ denote an attributed graph, where $\mathcal{V}=[n]$ and $\mathcal{E} \subseteq \mathcal{V} \times \mathcal{V}$ are the node set and the edge set respectively. $X \in \RR^{n\times d}$ denotes the node attributes with $d$-dimension. Further, we use $\mathcal{N}_v$ to represent the set of nodes in the direct neighborhood of node $v$, i.e., $\mathcal{N}_v = \{u: (u,v)\in \mathcal{E}\}$.
\end{definition}

\begin{definition}[$m$-hop Subgraph] Given a graph $\mathcal{G}$ and a node set of interest $Q$, let $\mathcal{G}_Q^m$ denote the $m$-hop neighboring subgraph w.r.t the set $Q$. $\mathcal{G}_Q^m$ is the induced subgraph of $\mathcal{G}$, of which the node set $\mathcal{V}^m_Q$ includes the set $Q$ and all the nodes in $\mathcal{G}$ whose shortest path distance to $Q$ is less than or equal to $m$. Its edge set is a subset of $\mathcal{E}$, where each edge has both endpoints in its node set $\mathcal{V}^m_Q$. The nodes in $\mathcal{V}^m_Q$ still carry the original node attributes if $\mathcal{G}$ is attributed.
\end{definition}

\subsection{Graph Learning Problems and Background}
Now, we formally formulate the GRL and SGRL problems.
\begin{definition}[Graph Representation Learning (\textbf{GRL})]
Given a graph $\mathcal{G}$ and a queried set of nodes $Q$, graph representation learning aims to learn a mapping from the graph-structured data to some predicting labels as $f(\mathcal{G},Q)\rightarrow y$, where the mapping $f(\mathcal{G},Q)$ may reflect structures and node attributes of $\mathcal{G}$ and their relation to $Q$.
\end{definition}

Next, we define SGRL where for a particular query $Q$, the predictions are made based on the local subgraph around $Q$.
\begin{definition}[Subgraph-based GRL (\textbf{SGRL})] \label{def:SGRL} Given a node set $Q$ over an ambient graph $\mathcal{G}$ and a positive integer $m$, SGRL is to learn the mapping to some labels, which takes the $m$-hop neighboring subgraph of $Q$ in $\mathcal{G}$ as the input $f(\mathcal{G}_Q^{m}, Q)\rightarrow y$. An SGRL task typically is given some labeled node set queries $\{(Q_i,y_i)\}_{i=1}^L$ for training and other unlabeled node set queries $\{Q_i\}_{i=L+1}^{L+N}$ for testing.
\end{definition}

We list a few important examples of SGRL tasks. \textbf{Link prediction} seeks to estimate the likelihood of a link between two endpoints in a given graph. Additionally, it can be generalized to predict the type of links, such as relation prediction for heterogeneous graphs. In this case, the set $Q$ corresponds to a pair of nodes. The network scientific community has identified the importance of leveraging the local induced subgraphs for link prediction~\cite{liben2007link}. For example, the number of common friends (shown as neighbors in a social network) implies how likely two individuals may become friends in the future. Another generalized form of link prediction is \textbf{higher-order pattern prediction}, where the set $Q$ consists of three or more nodes. The goal is to predict whether the set of nodes in $Q$ will foster a covered edge (termed hyperedge).

\textbf{Graph neural networks (\textbf{GNNs}).} Canonical GNNs associate each node $v$ with a vector representation $\vh$, which is learned and updated by aggregating messages from $v$'s neighbors, as \vspace{-2mm}
\[\vh_v^k=\text{UPDATE}\left(\vh_v^{k-1}, \text{AGGREGATE}\left(\{\vh_u^{k-1}|u\in \mathcal{N}_v\}\right)\right).\vspace{-2mm}\] 
Here, UPDATE is implemented by neural networks while AGGREGATE is a pooling operation invariant to the order of the neighbors. By unfolding the neighborhood around each node, the computation graph to get each node representation forms a tree structure. According to Def.~\ref{def:SGRL}, canonical GNNs seem also able to perform SGRL by encoding the local subtree rooted at each node into a node representation (See Fig.~\ref{fig:example}a). Nevertheless, by this way, each node representation only \textbf{separately} reflects the subgraph around each node but cannot \textbf{jointly} represent the subgraph around multiple nodes, which yields the problem in Fig.~\ref{fig:example}. However, the SGRL framework considered in this work is able to learn the representation of the joint subgraph around a queried node set.

\subsection{Other Related Works}
Without exception, previous works focus on improving the scalability of canonical GNNs and their system support, but some of their techniques inspire the design of \proj.

To overcome the memory bottleneck of GPU when processing large-scaled graphs, sub-sampling the graph structure is a widely adopted strategy. GraphSAGE~\cite{hamilton2017inductive} and VR-GCN~\cite{chen2017stochastic} use uniform sampling schema and variance reduction technique respectively to restrict the size of node neighbors; PIN-SAGE~\cite{ying2018graph} exploits Personalized PageRank (PPR) scores to sample neighbors. FastGCN~\cite{chen2018fastgcn} and ASGCN~\cite{huang2018adaptive} perform independent layer-wise node sampling to allow neighborhood sharing. Cluster-GCN~\cite{chiang2019cluster} and GraphSAINT~\cite{zeng2019graphsaint} study subgraph-based mini-batching approaches to reduce the size of training graphs. Note that the subgraphs in our setting are substantially different from theirs, since our subgraphs work as features for queries while their subgraphs are a compensatory choice to achieve better scalability.

Many works better the system support for GNNs. DGL~\cite{wang2019deep} and PyG~\citep{fey2019fast} are designed for scalable single-machine GNN training. Marius~\cite{mohoney2021marius} is proposed to efficiently learn large-scale graph embeddings on a single machine. There are several distributed systems dedicated to GNNs: AliGraph~\cite{yang2019aligraph} addresses the storage issue of applying GNNs on massive industrial graphs; AGL~\cite{zhang2020agl} employs a subgraph-based system for GRL; ROC~\cite{jia2020improving} builds a multi-GPU framework for deeper and larger GNN models; Dorylus~\citep{thorpe2021dorylus} designs a CPU-based distributed system for GNN training. $\text{G}^3$~\cite{liu2020g3} speedups GNN training via supporting parallel graph-structured operations. \citet{zhou2021accelerating} uses feature dimension pruning to accelerate large-scale GNN inference. However, all these systems only support canonical GNNs so they all suffer from the intrinsic modeling limitations of GNNs.
\section{The Architecture of \proj} \label{sec:arch}
\begin{figure}[tp]
\centering
\includegraphics[width=0.99\columnwidth]{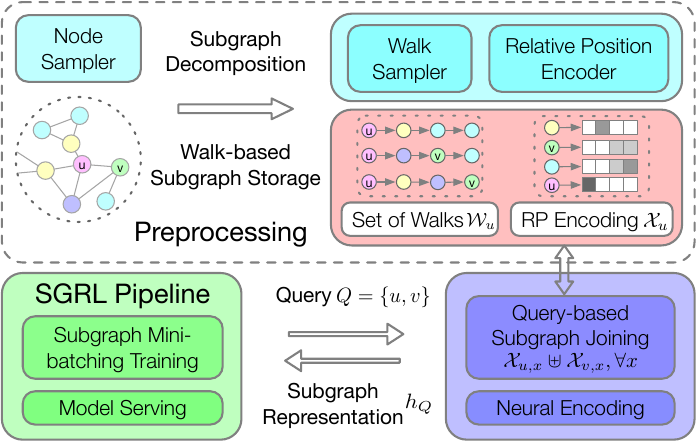}
\caption{The System Architecture of Subgraph-based Graph Representation Learning Framework (\proj).}
\label{fig:sgrl}
\end{figure}

In this section, we first give an overview of the \proj framework as shown in Fig. \ref{fig:sgrl}. Then, we focus on the design and the implementation of three modules: Walk Sampler \& Relative Position Encoder (Preprocessing), Walk-based Subgraph Storage, Query-based Subgraph Joining \& Neural Encoding. At last, we elaborate an efficient training pipeline with Subgraph Query Mini-batching.

\subsection{Overview}
Existing SGRL frameworks that extract a subgraph per query do not support efficient training and inference. $m$-hop subgraph extraction faces the size ``explosion'' issue as many nodes have significantly large degrees in real-world networks. Moreover, subgraphs of different sizes cause workload fluctuation, hindering load balancing and memory management.

Subgraph extraction can be replaced with efficient walk-based sampling, which sidesteps all above issues via regulating the number and the length of sampled walks. The number and the length of these walks are small constants, so the space and time complexity here is only linear w.r.t the number of nodes. Specifically, during preprocessing, \proj reduces the subgraph around each node in a given graph to a set of random walks originated from it. To compensate for the loss of structural information  after breaking subgraphs into walks, an intra-node distance feature termed relative positional encoding (RPE) is proposed, which enables locating each node in the sampled subgraph. The collected set of walks paired with its RPEs is hosted in the walk-based subgraph storage, with a dedicated data structure designed to support rapid and intensive access. The preprocessing flow is presented in the upper part of Fig. \ref{fig:sgrl}.

For training and testing, given a query (set of nodes), \proj employs \emph{subgraph joining} to implicitly construct a subgraph around the entire query in full parallel. First, all the walks originated from the queried node set are grouped. Then, the precomputed node-level RPEs are joined into query-level RPEs. \proj further adopts neural networks to encode the grouped walks paired with query-level RPEs, and makes final predictions based on the obtained subgraph representation. A mini-batching strategy is designed to maximize data reuse during training by exploiting the query overlaps.

\begin{figure}[tp]
\centering
\includegraphics[width=0.95\columnwidth]{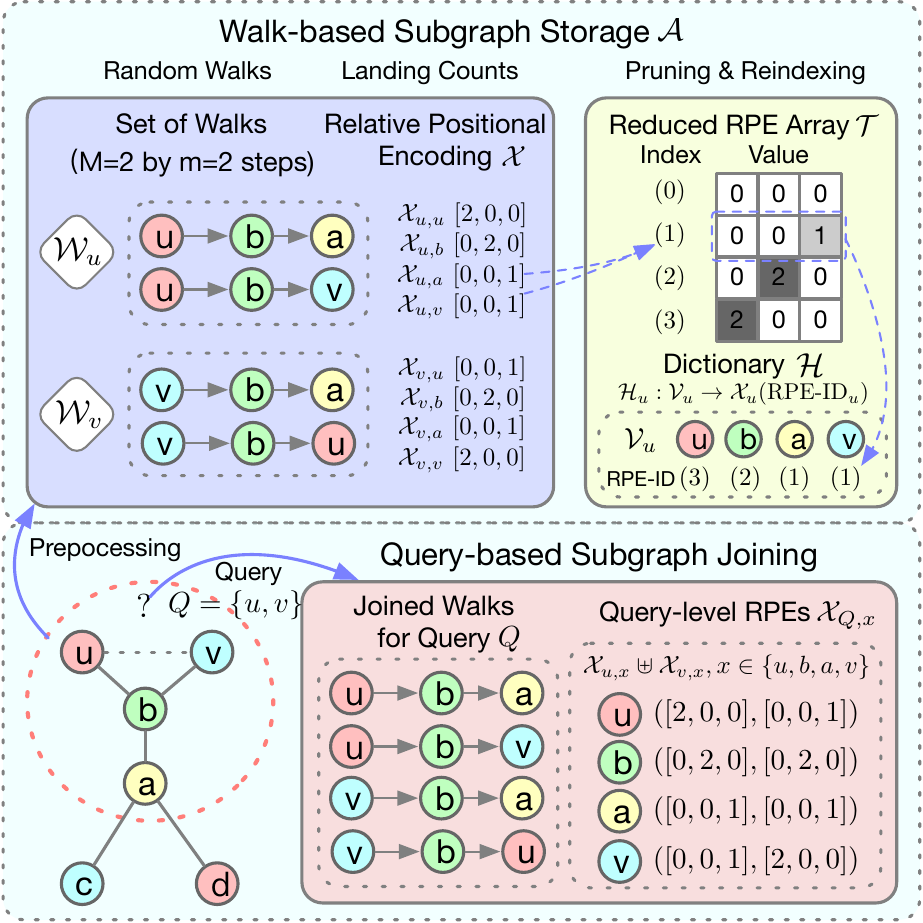}
\vspace{-2mm}
\caption{An Illustration of Joining RPE into Query-level RPEs with the Support of Walk-based Subgraph Storage.}
\vspace{-4mm}
\label{fig:gather}
\end{figure}

\subsection{Preprocessing - Walk Sampling \& Encoding} \label{sec:sampling}
The bottleneck of current SGRL frameworks is how to cheaply acquire the $m$-hop neighbors for each queried set of nodes. \proj proposes to decompose the $m$-hop subgraph into a set of $m$-length walks that start from the queried set of nodes. As the walks are regular, their storage and access are extremely efficient. This also resolves the computational problem caused by the long-tailed distribution of node degrees. More importantly, the collected walks grouped by their starting nodes can be shared and reused among different queries. Our design decouples SGRL from redundant subgraph extraction and enables the reusability of preprocessed data. We summarize the preprocessing routines with the support of hash-indexed storage in Algorithm \ref{alg:pdf} and introduce the specifics next.

\textbf{Walk Sampling.} During preprocessing, \proj samples $M$-many $m$-step walks for every node in a given graph. As Fig. \ref{fig:gather} (upper left) shows, the sampled walks are grouped in a set $\mathcal{W}_u$, where $u$ denotes the starting node of these walks. Walk sampling can be easily divided into parallelizable pieces. The parallelization is implemented based on NumPy and OpenMP framework in C. Moreover, to further accelerate walk sampling, we use compressed sparse row (CSR) to represent the graph. The CSR format consists of two arrays, \texttt{idxptr} of length $|\mathcal{V}|+1$ used to record the degrees of nodes, and \texttt{indices} of size $|\mathcal{E}|$, each row of which corresponds to the neighbor list per node. CSR allows intensive fast access to the neighbors of a node while keeping the memory cost low, which is vital for walk sampling in large-scale graphs.

\begin{algorithm}[tp]
\caption{\label{alg:pdf} Data Preprocessing in \proj} 
\KwIn{Graph $\mathcal{G}$; number of walks $M$; step of walks $m$}
\KwOut{Associative array $\mathcal{A}$, RPE array $\mathcal{T}$}
Initialize the array $\mathcal{A}$ and $\mathcal{T}$, the dictionary $\mathcal{H}$\\
\For{each node $u \in \mathcal{G}$}{
Run $M$ times $m$-step random walks on $\mathcal{G}$ as a set of walk $\mathcal{W}_u \in \ZZ^{M \times m}$\;
Add the key $\mathcal{V}_u = \text{set}(\mathcal{W}_u)$ to $\mathcal{H}_u$\;
Calculate RPE for $\forall x \in \mathcal{V}_u$, save the value $\mathcal{X}_{u,x}$ to $\mathcal{T}$, and write its index in $\mathcal{T}$ as $\text{RPE-ID}_{u,x}$ back to $\mathcal{H}_u(x)$\;
Insert $\{u: (\mathcal{W}_u, \mathcal{H}_u)\}$ to $\mathcal{A}$
}
Prune $\mathcal{T}$ and update the value of $\mathcal{H}$ by re-indexing.
\end{algorithm}
\begin{table}
\vspace{-6mm}
\end{table}

\textbf{Relative Positional Encoding (RPE).} 
Structural information gets lost after breaking subgraphs into walks. \proj compensates such loss via RPE to locate the relative position of a node in each sampled subgraph, which characterizes the structural contribution of the node to its corresponding subgraph. 

For each set of walks $\mathcal{W}_u$, we first establish a set $\mathcal{V}_u$ that contains distinct nodes appearing in $\mathcal{W}_u$. Define \emph{node-level} RPE $\mathcal{X}_{u}:\mathcal{V}_u\rightarrow \RR^{m+1}$ as follows: for each node $x \in \mathcal{V}_u$, a vector $\mathcal{X}_{u,x} \in \RR^{m+1}$ is assigned, where $\mathcal{X}_{u,x}[i]$ is the landing counts of node $x$ at position $i$ in all walks of $\mathcal{W}_u$. In \proj, RPE can be computed on the fly as walks get sampled, thus resulting in nearly zero extra computational cost. The set of walks $\mathcal{W}_u$ paired with the RPE $\mathcal{X}_u$ essentially characterize a sub-sampled subgraph around the node $u$. Next, we present a dedicated data structure to host $\mathcal{W}_u$ and $\mathcal{X}_u$ altogether.

\vspace{-1mm}
\subsection{Walk-based Subgraph Storage} \label{sec:storage}
It is easy to manage the collected set of walks due to its regularity. An $m*M$-sized chunk is allocated to each set of walks, which assists to speed up data fetching. How to organize node-level RPE presents a real challenge because the cardinality of the set $|\mathcal{V}_u|$ varies from node to node. One na\"ive way to avoid such irregularity is to directly scatter these RPEs back to nodes in previously collected walks. But, this gives an $m*M*(m+1)$ tensor, resulting in an unrealizable memory need. Moreover, it loses track of node IDs in walks that are needed for joining subgraphs later.

We use an associative array $\mathcal{A}$ to organize all walk-based subgraphs as shown in the upper part of Fig. \ref{fig:gather}. For each node $u\in \mathcal{V}$, its corresponding entry in $\mathcal{A}$ is a node-level subgraph formed as a tuple $(\mathcal{W}_u, \mathcal{H}_u)$. Here, $\mathcal{W}_u$ is a set of  walks starting from $u$, and $\mathcal{H}_u$ is a dictionary that maps the unique node set $\mathcal{V}_{u}$ of $\mathcal{W}_u$ to its corresponding node-level RPE $\mathcal{X}_{u}$. The use of dictionary resolves irregularities in $\mathcal{V}_{u}$ mentioned above, while maintaining the connection between node IDs and their RPEs. In addition, array $\mathcal{T}$ is introduced to store RPE values centrally, rather than scattered across dictionaries. As Fig. \ref{fig:gather} (upper right) shows, the value of $\mathcal{H}_u(x)$ is now replaced with the index of the RPE value $\mathcal{X}_{u,x}$ stored in $\mathcal{T}$ accordingly, noted as $\text{RPE-ID}_{u,x}$. This design overall guarantees the access of RPE in $O(1)$ time.

The above $\mathcal{A}$ and $\mathcal{H}_u$ are built on top of \texttt{uthash}'s macros \footnote{\url{https://troydhanson.github.io/uthash/}}, with extended support for arbitrary insertions and deletions of key–value pairs. It offers data access and search in $O(1)$ time on average, which is about as good as the direct address table but greatly reduces the space wastage. In particular, it has no dependency or need for communication between multiple hash queries, thus can be pleasingly executed in parallel. Both $\mathcal{A}$ and $\mathcal{H}_u$ are stored in RAM on the CPU side. As we observed in Fig. \ref{fig:gather}, there are many repeated RPE values. Once all nodes are sampled, the array $\mathcal{T}$ can be pruned to remove duplicates. $\text{RPE-ID}$s will be updated synchronously when $\mathcal{T}$ is reindexed. For example, both node $a$ and $v$ have the RPE value of $[0,0,1]$, whose index in $\mathcal{T}$ is (1) after pruning. Thus, both $\mathcal{H}_u(a)$ and $\mathcal{H}_u(v)$ are assigned to the new RPE-ID as (1). The shape of $\mathcal{T}$ is regular and its size is usually small after pruning, which can be fully loaded in GPU. In practice, we found that pining RPEs in GPU memory is critical, as it can significantly reduce the communication cost of moving data back and forth between RAM and SDRAM.

\vspace{-1mm}
\subsection{Query-based Subgraph Joining} \label{sec:joining}
The storage designed above records the downsampled subgraph around each node. As SGRL is mostly useful for making predictions over a set of nodes $Q$, here we further illustrate how to get the joined subgraph around all the nodes $u\in Q$.

The idea is to concatenate all set of walks $[...,\mathcal{W}_u,...]$ for $u\in Q$, since each set of walks $\mathcal{W}_u$ can be viewed as a subgraph around $u$. Besides, each node $x$ in the walks will be paired with a \emph{query-level} RPE $\mathcal{X}_{Q,x}$ that characterizes the relative position of node $x$ in the joint subgraph around the queried set $Q$. Specifically, $\mathcal{X}_{Q,x}$ is defined by joining all RPEs $\mathcal{X}_{u,x}$ for $u \in Q$, i.e., $\mathcal{X}_{Q,x}=\uplus_{u\in Q}\mathcal{X}_{u,x}(\triangleq[...,\mathcal{X}_{u,x},...])\in\RR^{(m+1)\times|Q|}$. There will be some $u\in Q$ such that $x\notin\mathcal{V}_u$, for which $\mathcal{X}_{u,x}$ is set to all zeros. Through this procedure, the joined subgraph with query-level RPEs is sent to GPU for representation learning and then model inference.

The data structure described in Sec. \ref{sec:storage} enables a highly parallel implementation of subgraph joining along with optimized memory management. On the CPU side, $\mathcal{X}_{Q,x}$ is not directly used to assemble walks. Instead, we use a query-level RPE-ID that joins node-level RPE indices in $\mathcal{T}$, i.e. use $\text{RPE-ID}_{Q,x} =[...,\text{RPE-ID}_{u,x},...]\in\RR^{|Q|}$ for $u \in Q$, which reduces the memory cost from $(m+1)*|Q|$ to $|Q|$. For instance, in Fig.~\ref{fig:gather} (bottom right), $\mathcal{X}_{Q,u}=([2,0,0],[0,0,1])$ can be substituted by $\text{RPE-ID}_{Q,u}=(3,1)$, as their RPE values locate at the entry (3) and (1) of $\mathcal{T}$. As follows, \proj pre-allocates an array with the fixed-size $[m*M*|Q|,|Q|]$, where $m*M*|Q|$ is the size of walks around $Q$. Then, \proj fills the index array with $\text{RPE-ID}_{u,x}$ by multithreads. Note that $\text{RPE-ID}_{u,x}$ can be rapidly retrieved via the dictionary operation $\mathcal{H}_u(x)$. Lastly, assembling RPE values to walks is performed on GPU via the indexing operation $\mathcal{X}_{u,x} = \mathcal{T}(\text{RPE-ID}_{u,x})$, where $\mathcal{T}$ is pinned in GPU memory earlier. \proj incorporates a Python/C hybrid API for subgraph joining, building on top of NumPy, PyTorch, OpenMP and \texttt{uthash}.

Some remarks can be made here. First, the above algorithm contains some redundancy to compute the query-level RPE-ID for the nodes that appear multiple times in the walks. In practice, we find that about half of the nodes appear only once, thus doubling the computation time at most. To avoid such redundancy, one can first compute the set union $\mathcal{V}_{Q} =\cup_{u\in Q}\mathcal{V}_u$, and then compute the query-level RPE-ID by traversing all nodes in $\mathcal{V}_{Q}$. However, parallel set union is difficult to implement efficiently. When multithreading is enabled, we observe a significant increase in the efficiency of \proj, as opposed to the union operation. Also, by dynamically adjusting the number of threads, the workload between CPU and GPU can be well balanced. Second, we have empirically found that using RPE-ID instead of RPE to assemble walks provides an observable performance boost (speed up by $2\times$ or more), otherwise data communication between CPU and GPU would the main bottleneck.

\vspace{-1mm}
\subsection{Neural Encoding}
After subgraph joining for each query, the obtained subgraph is represented by a concatenated set of walks on which nodes are paired with query-level RPEs (See Fig.~\ref{fig:gather}). Next, we introduce neural networks to encode these walks into a subgraph representation $h_Q$. 

Due to its regularity, any sequential models, e.g., MLP, CNN, RNN, and transformers can be adopted for sampled walks. We test RNN and MLP for neural encoding, both of which achieve similar results. Next, we take the RNN as an example. We encode each walk $W =(w_0,w_1,...,w_m) \in \mathcal{W}$ as $\text{enc}(W) = \text{RNN}(\{f\left(\mathcal{X}_{Q,w_i}\right)\}_{i=0,1,\dots,m})$, where $w_i$'s denote the node at step $i$ in one sampled walk. Here, $f$ is to encode the query-level RPE. Node or edge attributes for each step $w_k \in W$ can be supported by attaching those attributes after its RPE. To obtain the final subgraph representation of $Q$, we aggregate the encoding of all the associated walks through a mean pooling, i.e., $h_{Q}=\text{mean}(\{\text{enc}(W)|W\;\text{starts from some $u\in Q$}\}).$
In the end, a two-layer classifier is used to make prediction by taking $h_{Q}$ as input. In our experiments, all the tasks can be formulated as binary classification, and thus we adopt Binary Cross Entropy as the loss function.

\begin{algorithm}[tp]
\caption{\label{alg:sgrl} The Training Pipeline of \proj } 
\KwIn{A graph $\mathcal{G}$, a set of training queries $\{(Q_i, y_i)\}$, batch capacity $B_1$, batch size $B_2$}  
\KwOut{A Neural Network for Neural Encoding $\text{NN}(\cdot)$}
Prepare the collection of set of walks $\mathcal{W}$ and RPEs $\mathcal{X}$\\
\For{$iter = 1,\dots,max\_iter$}{
Initialize the set $\mathcal{Q}=\emptyset$ to track reached queries\;
Randomly choose a seed-set of nodes $\bar{\mathcal{V}}$ from $\cup Q$\;
Run breath-first search to expand $\bar{\mathcal{V}}$ and $\mathcal{Q}$ until $|\bar{\mathcal{V}}|=B_1$ or $|\mathcal{Q}|=B_2$\;
Generate negative training queries (if not given) for a mini-batch and put them into $\mathcal{Q}$\;
Perform subgraph joining for queries in $\mathcal{Q}$\;
Encode the concatenated walks by $\text{NN}(\cdot)$ to get the subgraph representation $h_{Q}$ for each query\;
Use backpropagation \cite{rumelhart1986learning} to optimize model parameters.
}
\end{algorithm}
\begin{table}
\vspace{-8mm}
\end{table}

\vspace{-1mm}
\subsection{The Training and Serving Pipelines}
\proj organically incorporates the storage designed in Sec.~\ref{sec:storage} and the subgraph-joining operation described in Sec.~\ref{sec:joining} to achieve efficient training and model serving.  

Subgraph queries $Q$'s are sets of nodes, which often come from a common ambient on a large graph. There might be many overlaps between different queries and their $m$-hop induced subgraphs. If the queried subgraphs are known in prior, we may put these queries with high node overlap into the same batch to improve data reuse. Here, queries of each given task are assumed to have the same size, e.g. $|Q|=2$ for link prediction. In practice, test queries are usually given online while the training ones can be prepared in advance. Hence, we propose to accelerate the training pipeline by mini-batching the overlapping queries. Practitioners can choose the appropriate pipeline according to the specific situation. Algorithm \ref{alg:sgrl} summarizes the overall training procedure of \proj.

\textbf{Mini-batching for Training.} We first randomly sample a seed-set of nodes $\bar{\mathcal{V}}$ from the union of queried node sets $\cup Q$. Then, we run breadth-first search (BFS) to expand the seed-set $\bar{\mathcal{V}}$. Neighbor fetching of the BFS here is based on the grouped queries instead of the original graph: a neighbor of node $u$ is defined as the node that shares at least one query with it. During BFS, the reached queries will be added to a set $\mathcal{Q}$. The expansion stops once the size of either the seed-set $\bar{\mathcal{V}}$ or the mini-batch $\mathcal{Q}$ reaches some pre-defined limits. Since the data structure for each query in \proj after subgraph joining is regular, it is easy to decide the size limits of seed-set and mini-batch based on resource availability (i.e. GPU memory). In practice, this BFS procedure improves reusability of data within each mini-batch, and may significantly decrease the communication cost between CPU and GPU. If the training set only contains positive queries (often in link/motif prediction tasks), we design an efficient sampling strategy for negative queries by the same principle that randomly pairs them within the same batch.
\section{Evaluation\label{sec:exp}}
In this section, we aim to evaluate the following points:
\begin{itemize}
    \item Regarding prediction performance, can \proj outperform state-of-the-art SGRL models? Can \proj significantly outperform canonical GNNs and transductive graph embedding methods due to the claimed benefit of SGRL?
    \item Regarding runtime, can \proj significantly outperform state-of-the-art SGRL models? Can \proj achieve runtime performance comparable to canonical GNNs? Previous SGRL models are typically much slower than canonical GNNs.
    \item How about the parameter sensitivity of \proj? How do the parameters $m$ and $M$ impact the overall performance?
    \item How is the parallel design of \proj performing and scaling?
\end{itemize}

\begin{table}[tp]
\centering
\caption{Summary Statistics for Evaluation Datasets.}
\vspace{-3mm}
\label{tab:data}
\begin{center}
  \resizebox{0.48\textwidth}{!}{
  \begin{tabular}{lcccccccc}
    \toprule
    \textbf{Dataset}&\textbf{Type}&\textbf{\#Nodes}&\textbf{\#Edges}\\
    \midrule
    \texttt{citation2} & Homo. & 2,927,963 & 30,561,187 \\
    \texttt{collab} & Homo. & 235,868 & 1,285,465 \\
    \texttt{ppa} & Homo. & 576,289 &  30,326,273 \\
    \texttt{ogb-mag} & Hetero. & \begin{tabular}[c]{@{}l@{}}Paper(P): 736,389\\ Author(A): 1,134,649\end{tabular} & \begin{tabular}[c]{@{}l@{}}P-A: 7,145,660\\ P-P: 5,416,271\end{tabular} \\
    \texttt{tags-math} & Higher. & 1,629 & \begin{tabular}[c]{@{}l@{}}91,685 (projected) \\ 822,059 (hyperedges) \end{tabular}\\
    \begin{tabular}[l]{@{}l@{}} \texttt{DBLP-} \\ \texttt{coauthor} \end{tabular} & Higher. & 1,924,991  & \begin{tabular}[c]{@{}l@{}}7,904,336 (projected) \\ 3,700,067 (hyperedges) \end{tabular}\\
  \bottomrule 
\end{tabular}}
\end{center}
\vspace{-3mm}
\end{table}

\subsection{Evaluation Setup}
We conduct extensive experiments to evaluate the proposed framework with three kinds of graphs (homogeneous, heterogeneous, and higher-order homogeneous) on three corresponding types of tasks, namely, link prediction, relation prediction and higher-order pattern prediction. Homogeneous graphs are the graphs without node/link types. Heterogeneous graphs include node/link types. Higher-order graphs contain higher-order links that may connect more than 2 nodes. The dataset statistics are summarized in Table \ref{tab:data}, most of which are larger than the datasets used in~\cite{zhang2021grain,zhou2021accelerating}, not to mention that our node-set prediction task is much more complex than the node classification task considered in the previous works.

\textbf{Open Graph Benchmark (OGB).} We use three link prediction and one relation prediction datasets \citep{hu2020ogb}: \texttt{ppa} - a protein interaction network, \texttt{collab} - a collaboration network, and \texttt{citation2} - a citation network; and one heterogeneous network \texttt{ogb-mag}, which contains four types of nodes (paper, author, institution and field) and their relations extracted from MAG \cite{wang2020microsoft}.

\textbf{Higher-order Graph Dataset.} \texttt{DBLP-coauthor} is a temporal higher-order network that records co-authorship of papers as timestamped higher-order links. \texttt{tags-math} contains sets of tags that are applied to questions on the website \url{math.stackexchange.com} as higher-order links. For the two higher-order graphs, \proj and all the baselines will treat them as standard graphs by projecting higher-order links into cliques. However, the training and test queries are generated based on higher-order links detailed next.

\textbf{Settings.} For \emph{Link Prediction}, we follow the data split as OGB requires to isolating the validation and test links (queries) from the graphs. For \emph{Relation Prediction}, the relations of paper-author (P-A) and paper-citation (P-P) are selected. The dataset is split based on timestamps. 0.5\% of existing edges of each target relation type are selected from \texttt{ogb-mag}. For each paper, two authors/citations are picked from its P-A/P-P relations respectively, one for validation and the other for testing. The remaining links are used for training. For \emph{Higher-order Pattern Prediction}, we focus on predicting whether two nodes will be connected to a third node concurrently via a higher-order link in the future. Specifically, positive queries are node triplets, where two nodes are linked before the timestamp $t$ and the third node establishes connection to the pair via a higher-order link after $t$. The split ratio of positive node triplets is 60/20/20 for training/validation/testing. For \emph{Relation Prediction} and \emph{Higher-order Pattern Prediction}, each positive query is paired with 1000 randomly sampled negative queries (except \texttt{tags-math} uses 100) in testing. For fair comparison, all baselines are tested with the same set of negative queries sampled individually for each dataset. All experiments are run 10 times independently, and we report the mean performance and standard deviations.

\begin{table}[tp]
    \centering
    \caption{Comparison of SGRL Methods for Subgraph Sampling. Suppose using $O(|\mathcal{E}|)$ many queries and $S$ to denote the average size of sampled subgraphs. The wall-clock time is measured on \texttt{citation2} test set with $p=16$ threads.}
    \vspace{-2mm}
    \label{tab:prep}
    \resizebox{0.48\textwidth}{!}{
    \begin{tabular}{c|c|c|c}
        \toprule 
        \textbf{Methods} & SEAL (1-hop)~\cite{zhang2018link,zhang2021labeling} & DE-GNN~\cite{li2020distance} & \proj \\
        \hline 
        Time Complexity  &  $O(S|\mathcal{E}|)$ & $O(S|\mathcal{E}|)$ & $O(\frac{mM}{p} \cdot |\mathcal{V}|)$\\
        Wall Time & 36h & > 1 month  & 26s\\
    \bottomrule
    \end{tabular}}
    \vspace{-5mm}
\end{table}

\textbf{Baselines.} We consider three classes of baselines. \textit{Graph Embedding methods} for transductive learning: Node2vec \cite{grover2016node2vec} and DeepWalk \cite{perozzi2014deepwalk}, which learns a single embedding for each node and may suffer from the information over-squashing issue; \textit{Canonical GNNs}: GCN \cite{kipf2016semi}, GraphSAGE \cite{hamilton2017inductive}, GraphSAINT \cite{zeng2019graphsaint}, Cluster-GCN \cite{chiang2019cluster}, Relational GCN (R-GCN) \cite{schlichtkrull2018modeling}, Relation-aware Heterogeneous Graph Neural Network (R-HGNN) \cite{yu2021heterogeneous}; \textit{SGRL models}: SEAL\cite{zhang2018link,zhang2021labeling}, DE-GNN\cite{li2020distance}. SEAL supports both offline and online subgraph extraction per query. However, it takes SEAL 2+ hours and 102GB RAM to offline extract 2\% training subgraphs on \texttt{citation2}. Thus, we only keep the online setting for SEAL. DE-GNN only supports offline subgraph extraction. Table \ref{tab:prep} compares subgraph sampling for different SGRL methods. We adopt official implementations of above baselines with tuned parameters that match reported results. \proj uses an 2-layer MLP for embeddings of RPEs and an 2-layer RNN to encode query-level joined walks. The obtained subgraph embeddings are fed into an MLP classifier for final prediction. Default training parameters are: learning rate \texttt{lr=1e-3} with early stopping of 5-epoch patience, dropout \texttt{p=0.1}, Adam \cite{kingma2014adam} as the optimizer, batch capacity $B_1=1500$, and batch size $B_2=32$. Hidden dimension $d$ and walk parameters $M,m$ are investigated in Sec. \ref{sec:mM}.

\textbf{Metric.}
The evaluation metrics include Hits@K and Mean Reciprocal Rank (MRR). Hit@K counts the percentage of positive samples ranked at the top-K place against all the negative ones. MRR firstly calculates the inverse of the ranking of the first correct prediction against the given number of paired negative samples, and then an average is taken over the total queries.

\textbf{Environment.} We use a server with four Intel Xeon Gold 6248R CPUs, 1TB DRAM, and eight NVIDIA RTX 6000 (24GB) GPUs. 

\newcolumntype{g}{>{\columncolor{Gray}}c}
\begin{table}[tp]
\caption{Results for Link Prediction, Relation Prediction, and Higher-order Pattern Prediction.}
\vspace{-4mm}
\label{tab:ogb}
\begin{center}
\resizebox{0.98\columnwidth}{!}{
  \begin{tabular}{lccc}
    \toprule
    \multirow{2}{*}{\textbf{Models}} & \texttt{citation2} & \texttt{collab} & \texttt{ppa} \\
    & MRR (\%) & Hits@50 (\%) & Hits@100 (\%) \\
    \midrule
    \textbf{Node2vec} & 61.28±0.15 & 47.54±0.78 & 18.05±0.52\\
    \textbf{DeepWalk} & 84.47±0.04 & 49.08±0.93 & 27.80±1.71\\
    \midrule
    \textbf{GCN} & 84.74±0.21 & 44.75±1.07 & 18.67±1.32 \\
    \textbf{SAGE} & 82.60±0.36 & 54.63±1.12 & 16.55±2.40 \\
    \textbf{Cluster-GCN} & 80.04±0.25 & 44.02±1.37 & 3.56±0.40 \\
    \textbf{GraphSAINT} & 79.85±0.40 & 53.12±0.52 & 3.83±1.33  \\
    \midrule
    \textbf{SEAL} & \underline{87.67±0.32} & \textbf{63.64±0.71} & \underline{48.80±3.16} \\
    \textbf{\proj} & \textbf{89.74±0.18} & \underline{63.34±0.52} & \textbf{53.23±1.03}  \\
  \bottomrule
\end{tabular}}
\label{tab:hop}
\resizebox{0.99\columnwidth}{!}{
\begin{tabular}{lcccc}
\toprule
    \multirow{2}{*}{\textbf{Models}} & \texttt{MAG(P-A)} & \texttt{MAG(P-P)} & \texttt{tags-math} & \texttt{DBLP-coauthor} \\
    &  MRR (\%) & MRR (\%)  & MRR (\%) & MRR (\%) \\
\midrule
    \textbf{GCN} & 39.43±0.29 & 57.43±0.30 & 51.64±0.27 & 37.95±2.59\\
    \textbf{SAGE} & 25.35±1.49 & 60.54±1.60 & 54.68±2.03 & 22.91±0.94\\
    \textbf{R-GCN} & 37.10±1.05 & 56.82±4.71 & - & -  \\
    \textbf{R-HGNN} & 33.41±2.47 & 45.91±3.28 & - & - \\
    \midrule
    \textbf{DE-GNN} & - & - & 36.67±1.59 & Timeout \\
    \textbf{\proj} & \textbf{45.33±2.94} & \textbf{82.47±0.26} & \textbf{71.86±2.15} & \textbf{97.66±2.89}\\
\bottomrule
\end{tabular}}
\end{center}
\vspace{-6mm}
\end{table}

\subsection{Prediction Performance Analysis}
Table \ref{tab:ogb} shows results of three prediction tasks. Apparently, for these three link prediction benchmarks, the performance of SGRL models is significantly better than transductive graph embedding models and canonical GNNs, particularly for the challenging tasks over \texttt{ppa} and \texttt{collab}. Within SGRL models, \proj sets two SOTA results on \texttt{ppa} and \texttt{citation2}, and gets comparable performance on \texttt{collab} against SEAL, which validates the modeling effectiveness of our proposed walk-based framework. For relation prediction and higher-order pattern prediction, we observe a large gap (up to 60\%) between canonical GNNs and \proj-based models, especially in higher-order cases. This again demonstrates the inherent modeling limitation of canonical GNNs to predict over a set of nodes. DE-GNN suffers from serious scalability issues when employing subgraph extraction for higher-order pattern prediction. Our best attempt is to deploy DE-GNN on \texttt{tags-math} by using 10\% training samples, while the other three graphs failed. DE-GNN spends more than 300 hours preprocessing just $5\%$ training queries of \texttt{DBLP-coauthor}.

\begin{table}[tp]
\caption{Breakdown of Runtime, Memory Consumption for Different Models on \texttt{citation2}, \texttt{collab} and \texttt{DBLP-coauthor}. Training time is calculated if no better validation result is observed in 3 consecutive epochs, which assumes the model has converged. Full-batch training models need NVIDIA A100 (48GB) GPUs, results of which are marked with *. Other models take less time on A100 than on RTX 6000.}
\vspace{-4mm}
\label{tab:cost_all}
\begin{center}
\resizebox{\columnwidth}{!}{
\begin{tabular}{l@{}l|c|rr|r|rr}
\toprule
& \multirow{2}{*}{\textbf{Models}} & \multicolumn{4}{c}{\textbf{Runtime (s)}}  & \multicolumn{2}{c}{\textbf{Memory (GB)}} \\ \cmidrule(r{0.5em}){3-8}
& & Prep. & Train & Inf. & Total & RAM & SDRAM \\
\midrule
\multirow{5}{*}{\rotatebox{90}{\texttt{citataion2}}}~~~& \ccolor\textbf{GCN *} & \ccolor17 & \ccolor16,835 & \ccolor32 & \ccolor16,884 & \ccolor9.5 & \ccolor37.55 \\
& \textbf{Cluster-GCN} & 197 & 2,663 & 82 & 2,942 & 18.3 & 14.07 \\
& \textbf{GraphSAINT} & 140 & 3,845 & 86 & 4,071 & 16.9 & 14.77 \\
& \textbf{SEAL (1-hop)} & 46 & 22,296 & 130,312 & 152,654 & 36.5 & 3.35 \\
& \textbf{\proj} & 31 & 2,096 & 7,959 & 10,086 & 15.2 & 4.50 \\
\midrule
\multirow{5}{*}{\rotatebox{90}{\texttt{collab}}}~~~& \textbf{GCN} & 6 & 840 & 0.1 & 846 & 3.2 & 5.17\\
& \textbf{Cluster-GCN} & 8 & 649 & 0.2 & 666 & 3.4 & 5.29 \\
& \textbf{GraphSAINT} & $<$1 & 6,746 & 0.2 & 6,747 & 3.2 & 6.58 \\
& \textbf{SEAL (1-hop)} & 10 & 7,675 & 37 & 7,722 & 15.4 & 6.97 \\
& \textbf{\proj} & $<$1 & 1,720 & 8 & 1,728 & 3.6 & 5.57\\
\midrule
\multirow{3}{*}{\rotatebox{90}{\texttt{DBLP}}}~~~& \ccolor \textbf{GCN *} & \ccolor- & \ccolor153 & \ccolor95 & \ccolor248 & \ccolor8.0 & \ccolor25.80\\
& \ccolor\textbf{SAGE *} & \ccolor- & \ccolor86 & \ccolor77 & \ccolor161 & \ccolor7.5 & 2\ccolor4.70 \\
& \textbf{\proj} & 10 & 430 & 1,667  & 2,107 & 8.6 & 8.61 \\
\bottomrule
\end{tabular}}
\end{center}
\vspace{-4mm}
\end{table}

\begin{figure}[tp]
\centering    
\includegraphics[width=0.92\columnwidth]{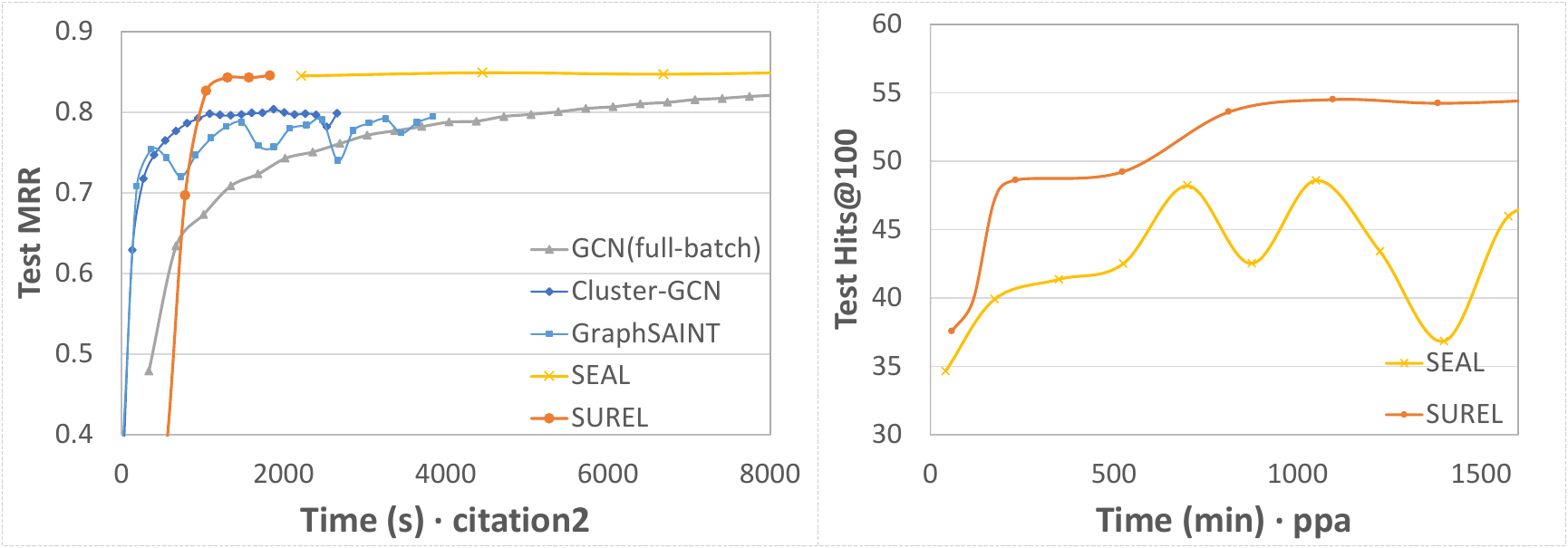}
\centering
\includegraphics[width=0.92\columnwidth]{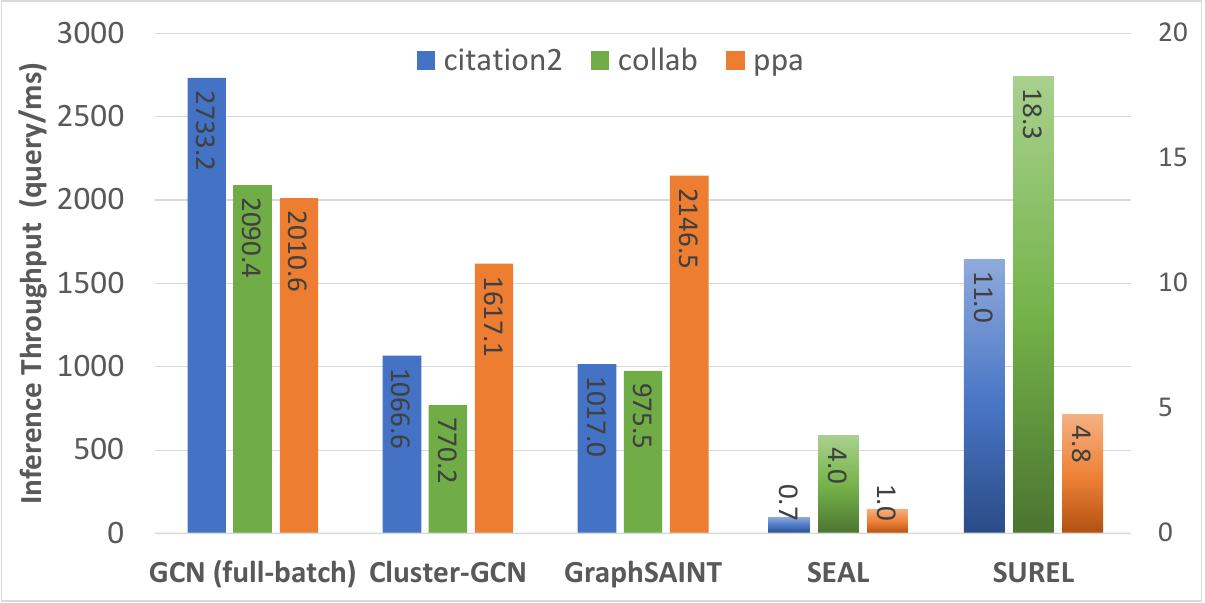}
\vspace{-2mm}
\caption{Performance Profiling of Training \& Inference (Up: Time-to-accuracy; Down: Inference Throughput).\label{fig:ttai}}
\vspace{-6mm}
\end{figure}

\begin{figure*}[htp]
\centering
\subfigure[Prediction Performance (m/M)\label{fig:mmp}]{
\begin{minipage}{0.235\textwidth}
\centering    
\includegraphics[width=4.3cm]{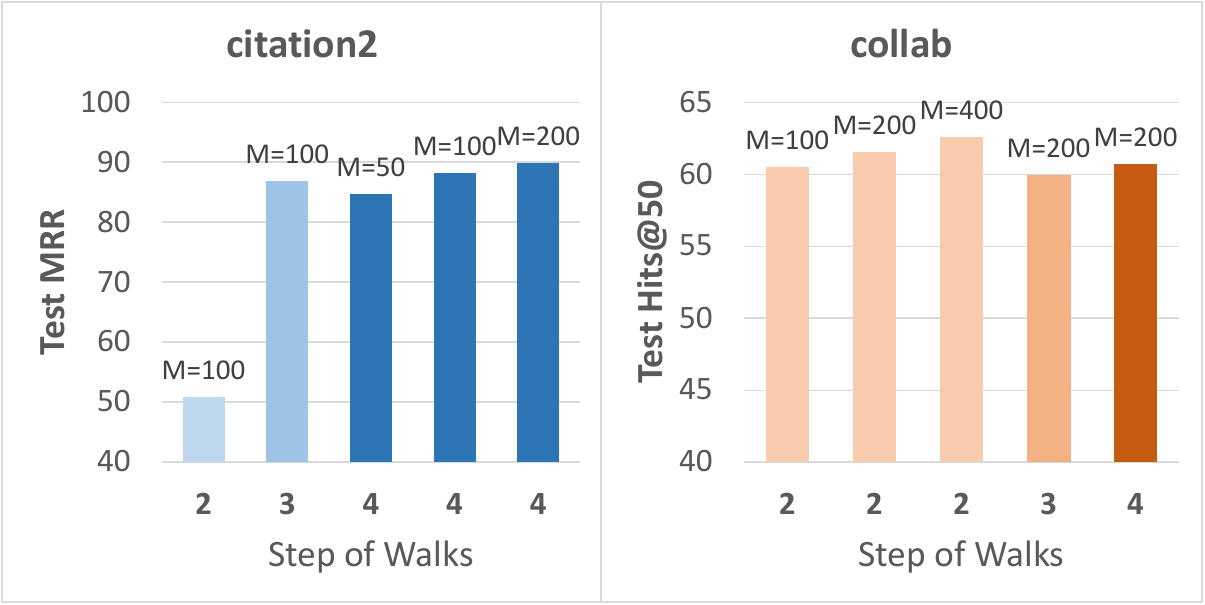}
\end{minipage}
}
\hfill
\subfigure[Training Time per Batch (2K queries) \label{fig:mmt}]{
\begin{minipage}{0.235\textwidth}
\centering
\includegraphics[width=4.3cm]{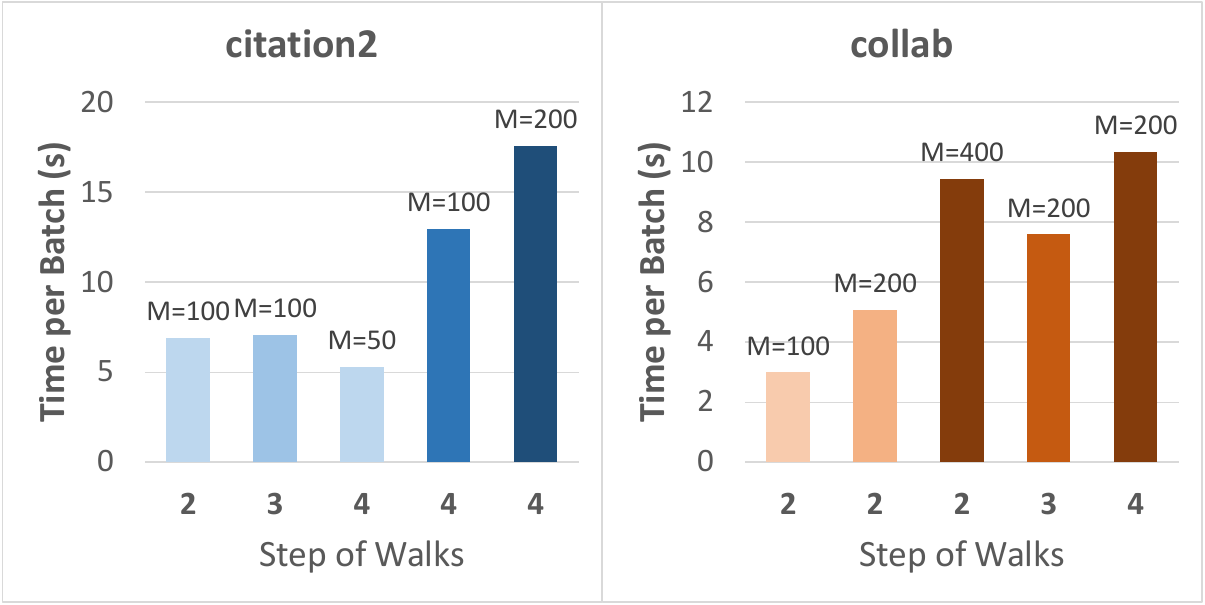}
\end{minipage}
}
\hfill
\subfigure[Inference Time \label{fig:mmf}]{
\begin{minipage}{0.235\textwidth}
\centering
\includegraphics[width=4.3cm]{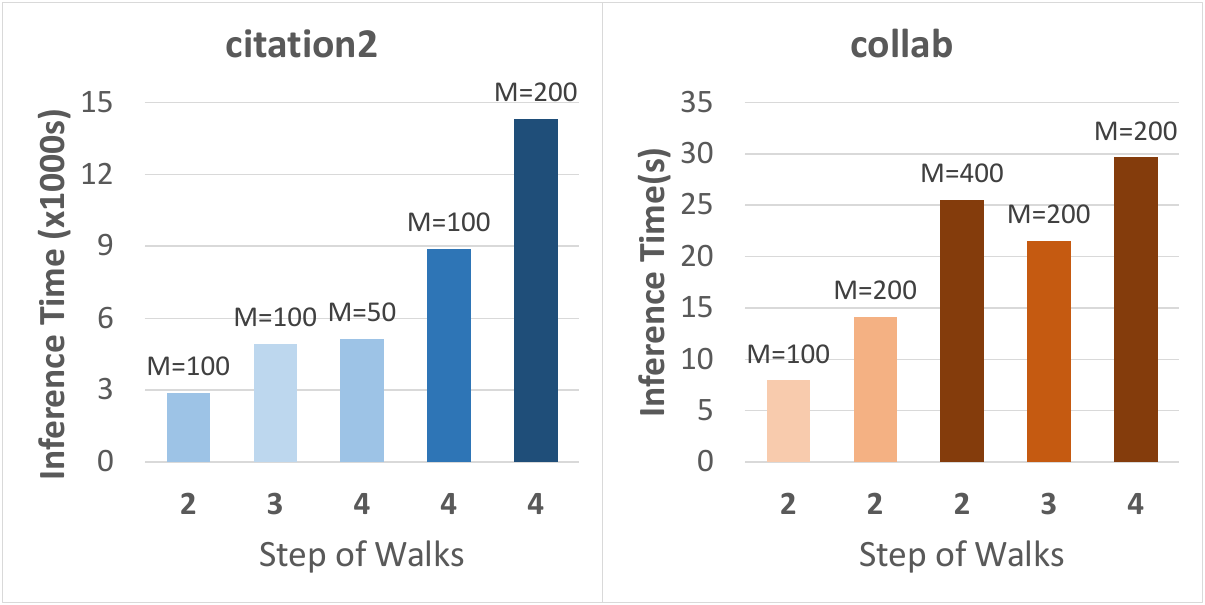}
\end{minipage}
}
\hfill
\subfigure[Prediction Performance (dim)\label{fig:dim}]{
\begin{minipage}{0.235\textwidth}
\centering    
\includegraphics[width=4.3cm]{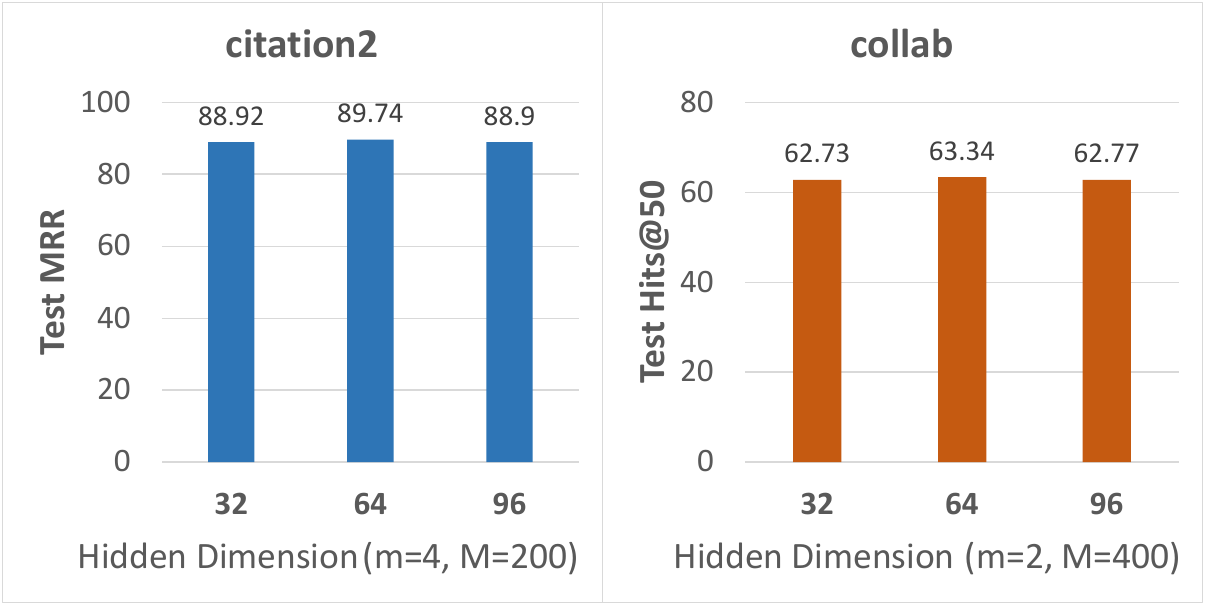}
\end{minipage}
}
\vspace{-4mm}
\caption{Hyper-parameter Analysis: the number of walks $M$, the step of walks $m$, and the hidden dimension $d$.}
\vspace{-4mm}
\label{fig:eff}
\end{figure*}

\subsection{Runtime and Memory Complexity Analysis}

Table \ref{tab:cost_all} reports the runtime, memory consumption comparison on a single machine (using one GPU) between canonical GNNs and SGRL models. \proj offers a reasonable total runtime on these benchmarks compared with canonical GNNs. Meanwhile, its preprocessing overhead is negligible as showed in Table \ref{tab:cost_all} under the term `\textbf{Prep.}', and the higher-order case can be efficiently handled as well. SEAL adopts online extraction, and thus the cost is not counted in preprocessing, while its training suffers from the computation bottleneck. DE-GNN uses offline extraction, and it takes 15+ hours and 98GB RAM to process training queries in \texttt{tags-math}, which is obviously incapable of scaling to \texttt{DBLP-coauthor} (so not present in Table \ref{tab:cost_all}). Overall, \proj substantially accelerates the subgraph extraction and makes it feasible for SGRL on large-scale graphs.

In terms of memory management, \proj achieves comparable RAM usage to canonical GNNs, because the number of walks $M$ and the steps $m$ are small constants in practice. The extra memory cost is linear in $|\mathcal{V}|$, so the total memory cost is still dominated by the original graph. However, SEAL induces much more RAM usage as it extracts subgraphs of long-tail sizes, and its total memory cost is often super-linear in $|\mathcal{V}|$. Both SEAL and \proj consume much less SDRAM because they do not need GPU to load large adjacency matrices and host node representations.

We further profile the training and inference performance, and present it in Fig. \ref{fig:ttai}. The upper half plots the time-to-accuracy comparison between canonical GNNs and SGRL models. Each dot indicates one training epoch for full-batch GCN, SEAL and \proj, 10 training epochs for Cluster-GCN and GraphSAINT. As it shows, both SEAL and \proj use 1-3 epochs to get good enough performance, and each epoch of \proj takes around 1/10 time of SEAL on \texttt{citation2}. The time per epoch of full-batch GCN is comparable with \proj, while Cluster-GCN and GraphSAINT are faster. However, these models generally take longer time to converge to even subpar performance. On \texttt{ppa}, the curve of SEAL is pretty oscillating, leading to longer convergence. \proj uses large $M$ and $m$ to achieve better and more stable performance on \texttt{ppa}, so the training time per epoch is comparable with SEAL. The training curves of canonical GNN baselines are not plotted for \texttt{ppa} because of their poor performance (See Table~\ref{tab:ogb}).

The bottom half of Fig. \ref{fig:ttai} provides the comparison of end-to-end inference throughput between two classes of models. Canonical GNNs offer rapid inference, since they generate node representations as the intermediate computation results that are shared across all queries. But as aforementioned, sharing node representations may over-squash useful information and degenerate performance as shown in Table~\ref{tab:ogb}. SEAL, as SGRL, achieves good prediction performance but its inference is extremely slow, because of subgraph extraction per query. \proj fundamentally solves this bottleneck by replacing the extraction with walk-based subgraph joining. It is $\textbf{4}-\textbf{16}\times$ faster than SEAL on inference for link prediction, and achieves even more speedup than DE-GNN in higher-order settings.

\begin{figure}[tp]
\hfill
\centering
\subfigure[Throughput\label{fig:tpre}]{
\begin{minipage}{0.22\textwidth}
\centering    
\includegraphics[width=3.6cm]{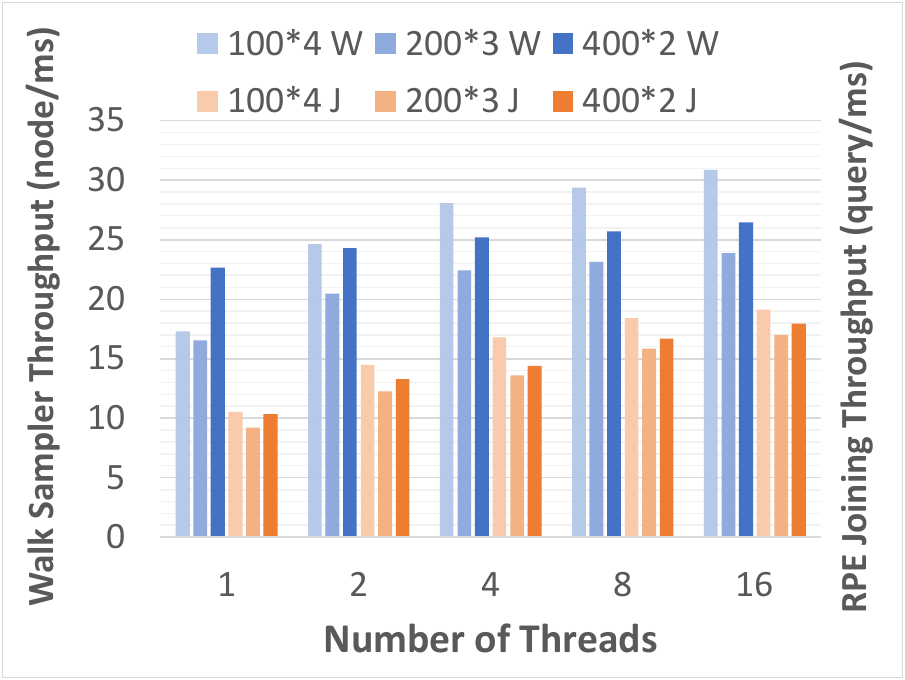}
\end{minipage}
}
\hfill
\subfigure[Runtime \label{fig:tall}]{
\begin{minipage}{0.23\textwidth}
\centering
\includegraphics[width=3.6cm]{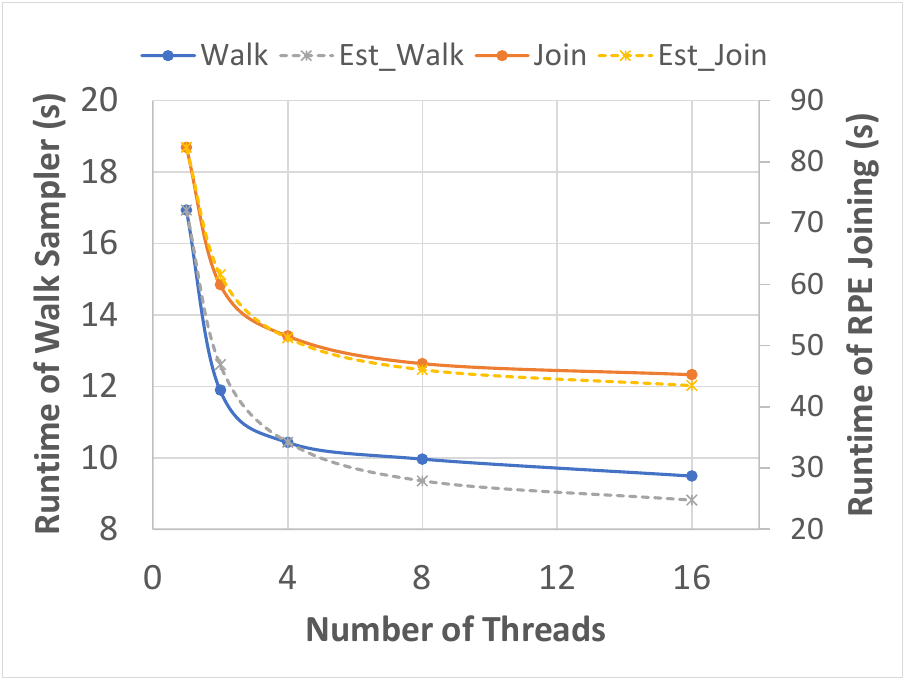}
\end{minipage}
}
\vspace{-4mm}
\caption{Performance Scaling of \proj (\textit{W}alk Sampler and Query-level RPE \textit{J}oining) vs Different Number of Threads.}
\vspace{-6mm}
\label{fig:scaling}
\end{figure}

\subsection{Significant Hyperparameter Analysis \label{sec:mM}}
The number $M$, the step $m$ of walks and the hidden dimension $d$ effect scalability and accuracy of \proj. To examine their impact, we evaluate \proj on \texttt{citation2}, a large sparse graph, and \texttt{collab}, a medium dense graph, for different values of $m$, $M$, and $d$. 

\textbf{Prediction Performance.}
Fig. \ref{fig:mmp} and \ref{fig:dim} show the prediction results. As expected, the performance consistently increases if we use a larger number of walks $M$. But for the step $m$, it is not always true that longer steps will guarantee better results, which depends on the specifics of the dataset. For instance, in network \texttt{citation2}, to accurately predict the link between two papers, more steps are needed as it would capture a larger group of papers which share similar semantics. While for \texttt{collab}, the case is different, as deeper walks would include more noise for predicting collaborations between two authors. In general, some small $m~(2\sim5)$ and $M~(50\sim400)$ ensure adequate performance. By adjusting $m$ and $M$, we can achieve the trade-off between accuracy and scalability, none of which is achievable through other SGRL models. Moreover, \proj is insensitive to the hidden dimension as shown by Fig.~\ref{fig:dim}.

\textbf{Training and Inference Time Cost.}
As Figs. \ref{fig:mmt} and \ref{fig:mmf} demonstrated, the time of walk sampling and subgraph joining is nearly linear w.r.t. the total number of walks ($m*M$) under the same number of threads (16 by default). Here, we do not regulate $M$ based on the degree of each node in a query, which may induce certain duplication in sampled walks originated from the nodes with small degrees. Using degree-adaptive $M$ is promising to further improve the scalability of \proj while keeping good prediction performance. We leave such investigation for future study.

\subsection{Performance Scaling}
To investigate the scaling performance of the parallel implementation, we examine the runtime of heavy operations in \proj by using different numbers of threads. Fig. \ref{fig:tpre} shows the throughput of walk sampler and query-level RPE joining on \texttt{citation2}. The runtime is also compared to the estimated runtime by Amdahl's law \cite{grama2003introduction} shown in Fig. \ref{fig:tall}: walk sampling and RPE joining are in good agreement with the expected speedup, thus implying well parallelized implementation.
\section{Conclusion}
We propose a novel computational paradigm, \proj for subgraph-based representation learning on large-scale graphs. \proj targets predicting relations over set of nodes. It decouples graph structures into sets of walks to avoid irregularities in subgraphs and enable reuse of intermediate results. It then applies walk-based subgraph joining paired with relative positional encoding for representation learning of queried node sets. Such design allows for full parallelization and significantly improves model scalability. \proj incorporates the principle of algorithm and system co-design that unlocks the full potential of learning on large-scale data with limited resources. To the best of our knowledge, this is the first work to study subgraph-based representation learning from the perspective of system scalability. Experiments also show that \proj achieves superior performance in both prediction and scalability on three different SGRL tasks over six large, real-world graph benchmarks.

\begin{acks}
We greatly thank all the reviewers for valuable feedback and actionable suggestions. H. Yin and P. Li are supported by the 2021 JPMorgan Faculty Award and the National Science Foundation (NSF) award HDR-2117997.
\end{acks}

\bibliographystyle{ACM-Reference-Format}
\bibliography{BIB/ref,BIB/gnnsystems,BIB/gnnmodel,BIB/background}

\appendix
\section{Notations}\label{apd:notation}
Frequently used symbols are summarized in Table \ref{tab:notation}. 

\begin{table}[tp]
\centering
\caption{Summary of Frequently Used Notations.}
\vspace{-3mm}
\label{tab:notation}
\begin{center}
  \begin{tabular}{lp{6.5cm}}
    \toprule
    \textbf{Symbol}&\textbf{Meaning}\\
    \midrule
    Q & a query (set of nodes), i.e. $Q=\{u,v,w\}$\\
    $\mathcal{Q}$ & a collection of queries, i.e. $Q\in \mathcal{Q}$\\
    $\mathcal{W}_u$ & the set of walks starting from node $u$\\
    $\mathcal{W}$ & a collection of walks, i.e. $W \in \mathcal{W}$\\
    $\mathcal{V}_u$ & the set of unique nodes appearing in $\mathcal{W}_u$\\
    $\mathcal{X}_u$ & the relative positional encoding (RPE) of nodes in $\mathcal{V}_u$ regarding their step occurrence in $\mathcal{W}_u$\\
    $\mathcal{X}_{u,v}$ & the RPE vector of node $v$ regarding the node $u$ (all zeros if $v \notin \mathcal{V}_u$)\\
    $\mathcal{T}$ & the RPE array with indices as RPE-IDs and entries as RPE vectors \\
    $\mathcal{H}_u$ & the dictionary with nodes in $\mathcal{V}_u$ as keys and RPEs $\mathcal{X}_u$ (or RPE-IDs) as values\\
    $\mathcal{A}$ & the associative array with nodes in $u \in \mathcal{V}$ as key and the tuple $(\mathcal{W}_u,\mathcal{X}_u)$ as entry\\
    $\uplus$ & the concatenation operation that joins all node-level RPEs, i.e. join $\mathcal{X}_{u,x}$ for $u \in Q$ as $[...,\mathcal{X}_{u,x},...]$.\\
    $\mathcal{X}_{Q,x}$ & the query-level RPE for node $x$, $\mathcal{X}_{Q,x}=\uplus_{u \in Q} \mathcal{X}_{u,x}$\\
  \bottomrule 
\end{tabular}
\end{center}
\vspace{-3mm}
\end{table}

\section{Further Analysis} \label{apd:analysis}
\begin{figure}[tp]
\includegraphics[width=0.495\columnwidth]{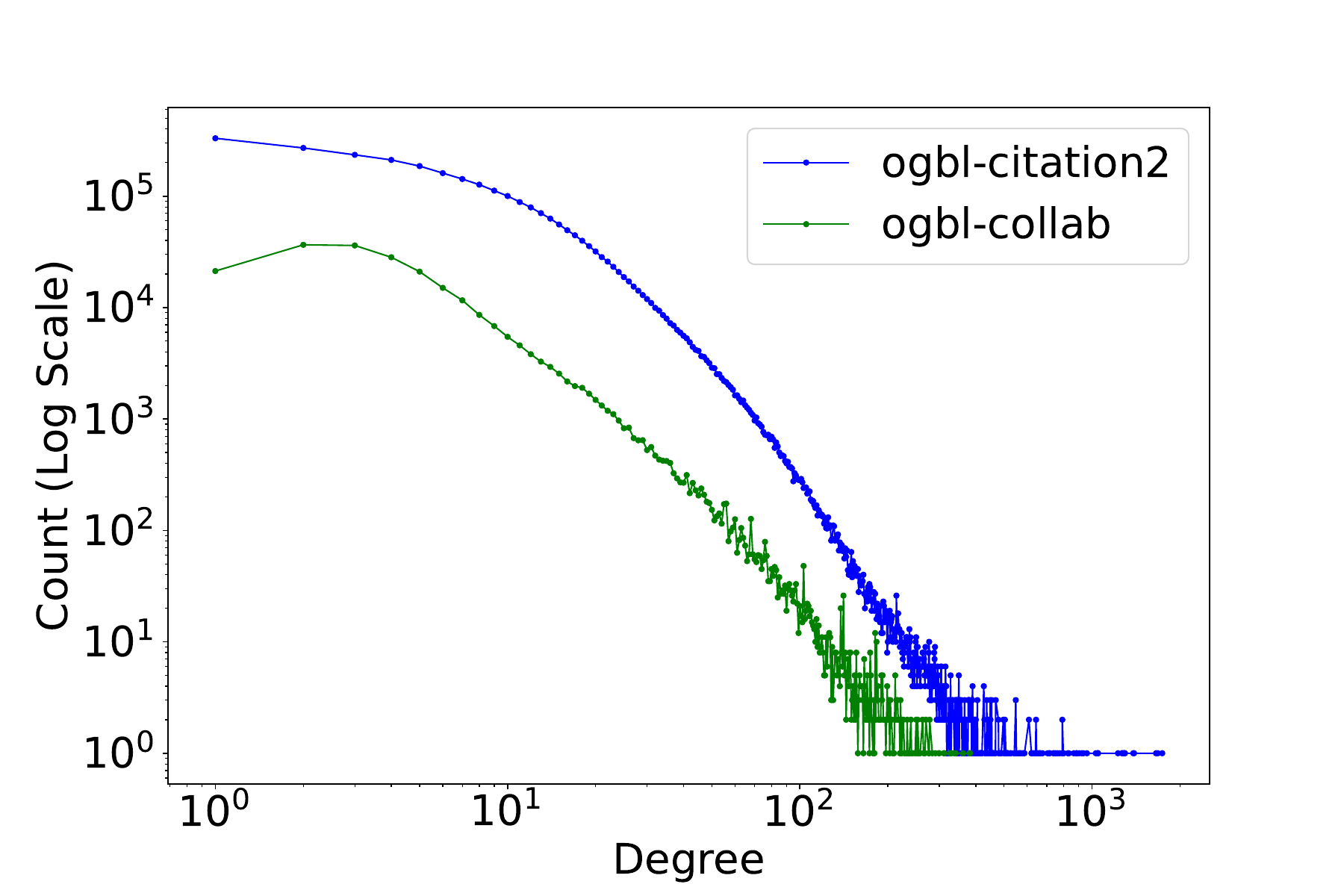}
\vspace{-2mm}
\includegraphics[width=0.495\columnwidth]{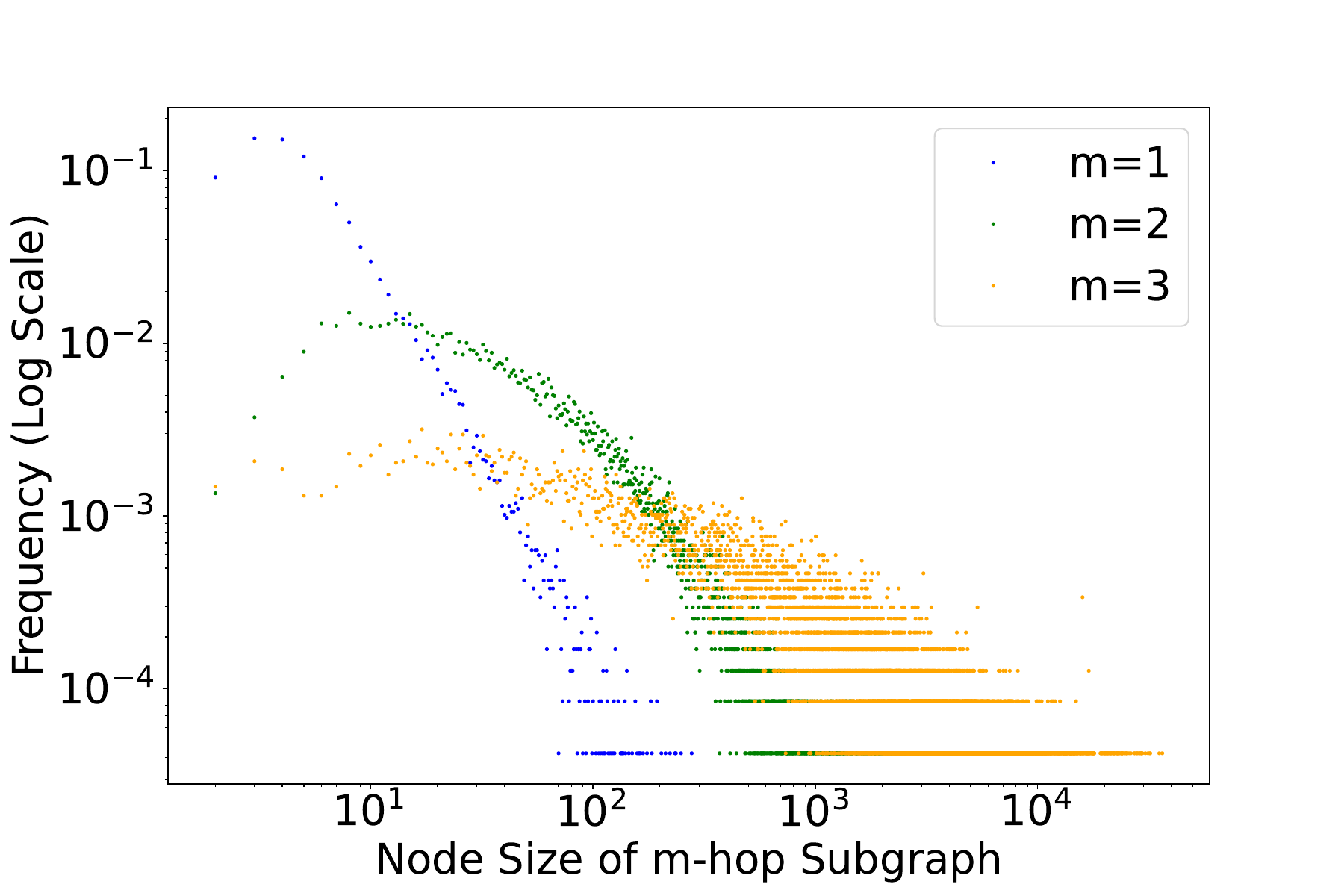}
\caption{Characteristics of Real-World Networks \label{fig:subgraph-size}}
\vspace{-3mm}
\end{figure}

Fig. \ref{fig:subgraph-size} shows the degree distribution and the node size of subgraph with respect to different numbers of hops in real-world networks \texttt{collab} and \texttt{citation2}. The node size of subgraphs dramatically increases when the number of hop $m\geq2$, because many nodes have significantly large degrees in real-world networks as Fig. \ref{fig:subgraph-size} LEFT illustrated. This leads to the size ``explosion'' issue for current SGRL models. Accordingly, most of SGRL models including SEAL~\cite{zhang2018link,zhang2021labeling} and DE-GNN~\cite{li2020distance} can only accommodate at most 1-hop neighbors over large networks to avoid the scalability crisis, which in return compromises their performance. \proj solves this issue by breaking the subgraph into regular walks, which enables it to reach up to $m$-hop neighbors via $m$-step random walks. The long-hop neighbors give extra information, which is beneficial for improving the prediction performance.

\section{Limitations of Canonical GNNs and More Illustration of the Algorithmic Insights of \proj}\label{adp:limitations}

Canonical GNNs are known to have several limitations, such as limited expressive power~\cite{xu2019powerful,morris2019weisfeiler}, feature oversmoothing~\cite{oono2019graph}, information over-squashing\cite{epasto2019single,alon2021bottleneck} and noise contaminating with a large receptive field~\cite{huang2020graph,zeng2021decoupling}.

\begin{figure}[tp]
\centering
\includegraphics[trim={2.5cm 22.1cm 8.0cm 2.4cm},clip,width=1.00\columnwidth]{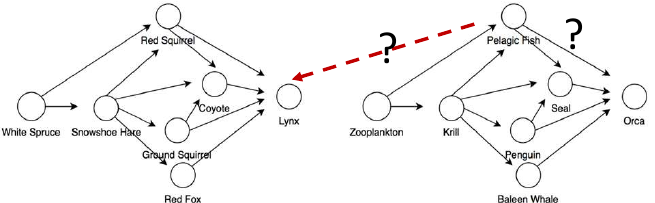}
\caption{A food web example shows two disconnected components - the boreal forest \citep{stenseth1997population} and the antarctic fauna \citep{bates2015construction}. The query here is which one is more likely the predator of Pelagic Fish, Lynx or Orca? Canonical GNNs cannot solve this query. \citet{srinivasan2019equivalence} explain this as the failure of GNNs to establish the correlation between the node representations.}
\label{fig:foodweb-de}
\vspace{-3mm}
\end{figure}

One of the biggest limitations is that canonical GNNs cannot distinguish the nodes that can be mapped to each other under some graph automorphism. For example, the nodes $w$ and $v$ in Fig.~\ref{fig:example} satisfy this property, and canonical GNNs will associate them with the same node representation. Another more practical example from a food web is shown in Fig.~\ref{fig:foodweb-de}. In fact, most large networks do not have non-trivial automorphism, and such nodes are not that common. However, the above issue of GNNs actually induces a more severe concern: the node representations learnt by GNNs cannot well capture the intra-node distance information, which is crucial to predict over a set of nodes.

Another issue of canonical GNNs is to over-squash information into a single node representation. Node representations can be viewed as intermediate computation results that are often used in several downstream tasks. However, a node representation, if it carries the information that is suitable for one task, may carry subpar information for another task.  

The third issue of canonical GNNs is the entanglement of the number of GNN layers and the range of the receptive field. In practice, when more complex and non-linear functions are to be approximated, one may want to add more layers to the neural network. However, GNNs adding more layers will also enlarge the receptive field that may introduce noise. 

\begin{figure*}[t]
\centering
\includegraphics[width=0.98\textwidth]{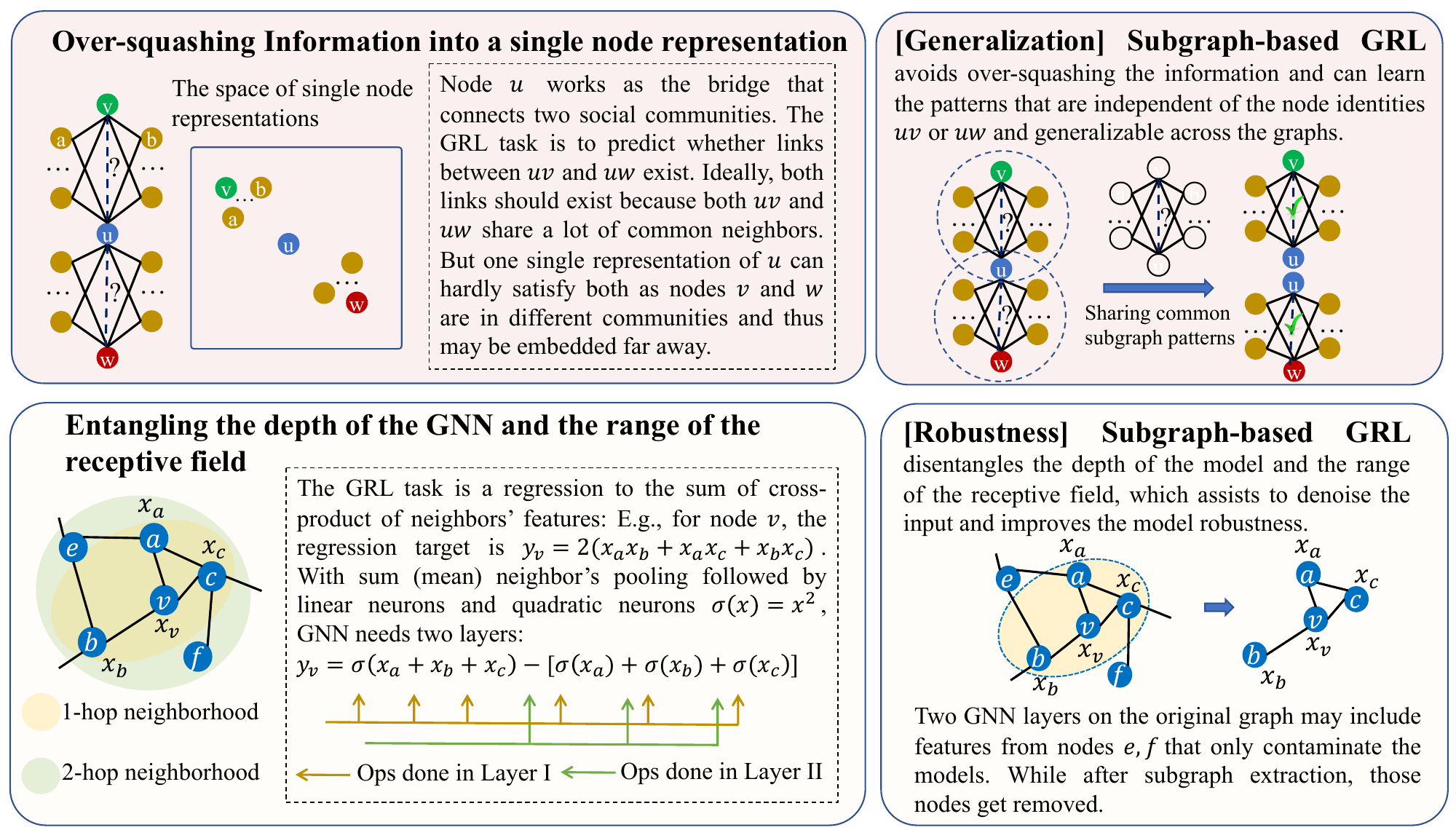}
\caption{LEFT: Two Limitations of Canonical-GNN-based GRL; RIGHT: How to Solve Them by Using Subgraph-based GRL.}
\label{fig:limitation}
\end{figure*}

The last two issues of GNNs are demonstrated in the left column of Fig.~\ref{fig:limitation}. SGRL methods can handle these two issues of GNNs in a simple way as illustrated in the right column of Fig.~\ref{fig:limitation}.

\begin{figure*}[t]
\centering
\includegraphics[width=0.98\textwidth]{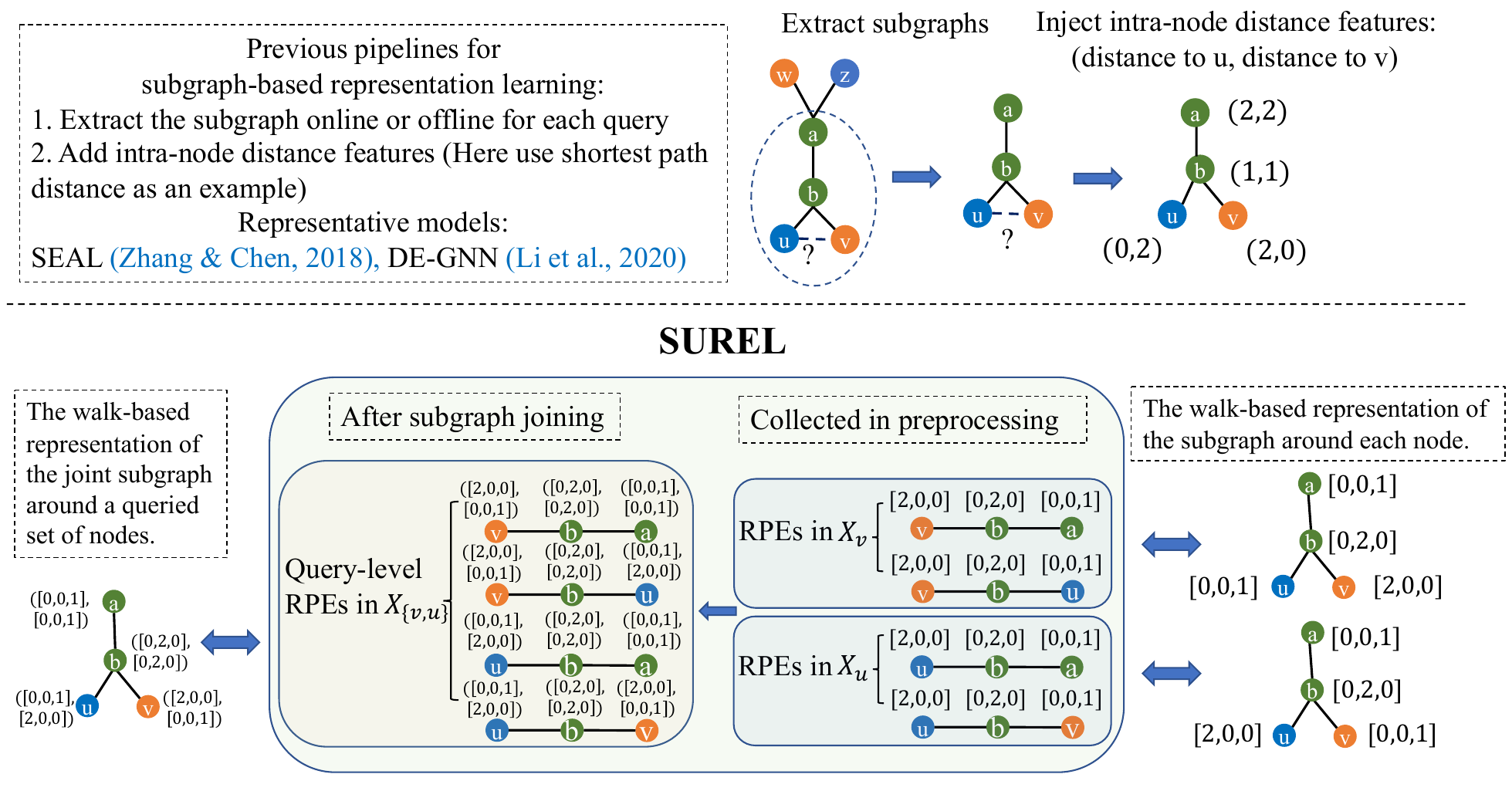}
\caption{How is \proj related to previous SGRL models? Previous SGRL models, such as SEAL~\citep{zhang2018link} and DE-GNN~\citep{li2020distance}, first extract the neighboring subgraphs according to the queries, and then attach distance features to the nodes. In \proj, all the subgraphs are represented by sets of walks sampled from the corresponding subgraphs. Given a query, the joint subgraph is represented by the concatenation of sets of walks. The query-level relative positional encoding (RPEs) is obtained by joining node-level RPEs that represent the intra-node distance features.}
\label{fig:alg-intuition}
\end{figure*}

Furthermore, to handle the challenge encountered in Fig.~\ref{fig:example}a, Fig.~\ref{fig:example}b indicates SGRL methods can be adopted, which is to consider the joint subgraph around the target nodes ($\{u,w\}$ or $\{u,v\}$) as a whole. Subgraph extraction can be mathematically viewed as adding an intra-node distance feature to each node $z$: suppose $Q=\{u,w\}$ is the queried node set; if the \textbf{distances} from $z$ to $u$ and $w$ are both less than or equal to 1, $z$ will be selected. Such distance features can also be used as extra node features directly attached to raw node features. \citet{li2020distance} have proved the effectiveness of using intra-node distance to better GNN expressive power. \proj is able to capture such intra-node distance by adopting relative positional encoding (RPE). Lastly, we illustrate how \proj is related to previous SGRL approaches in Fig.~\ref{fig:alg-intuition}.

\begin{table*}[tp]
\centering
\caption{Summary Statistics and Experimental Setup for Evaluation Datasets.}
\vspace{-2mm}
\label{tab:data-full}
  \resizebox{0.98\textwidth}{!}{
  \begin{tabular}{lcccccccc}
    \toprule
    \textbf{Dataset}&\textbf{Type}&\textbf{\#Nodes}&\textbf{\#Edges}&\textbf{Avg. Node Deg.}&\textbf{Density}&\textbf{Split Ratio}&\textbf{Split Type}&\textbf{Metric}  \\
    \midrule
    \texttt{citation2} & Homo. & 2,927,963 & 30,561,187 & 20.7 & 0.00036\% & 98/1/1 & Time & MRR\\
    \texttt{collab} & Homo. & 235,868 & 1,285,465 & 8.2 & 0.0046\% & 92/4/4 & Time & Hits@50\\
    \texttt{ppa} & Homo. & 576,289 &  30,326,273 & 73.7 & 0.018\% & 70/20/10 & Throughput & Hits@100\\
    \texttt{ogb-mag} & Hetero. & \begin{tabular}[c]{@{}l@{}}Paper(P): 736,389\\ Author(A): 1,134,649\end{tabular} & \begin{tabular}[c]{@{}l@{}}P-A: 7,145,660\\ P-P: 5,416,271\end{tabular} & 21.7 & N/A & 99/0.5/0.5 & Time & MRR\\
    \texttt{tags-math} & Higher. & 1,629 & \begin{tabular}[c]{@{}l@{}}91,685 (projected) \\ 822,059 (hyperedges) \end{tabular} & N/A & N/A & 60/20/20 & Time & MRR\\
    \texttt{DBLP-coauthor} & Higher. & 1,924,991  & \begin{tabular}[c]{@{}l@{}}7,904,336 (projected) \\ 3,700,067 (hyperedges) \end{tabular} & N/A & N/A & 60/20/20 & Time & MRR\\
  \bottomrule 
\end{tabular}}
\vspace{-2mm}
\end{table*}

\section{Additional Experimental Settings} \label{apd:exp}
\subsection{More Details about the Datasets}\label{apd:dataset}
The dataset statistics and experimental setup for evaluation are provided in Table \ref{tab:data-full}. We choose OGB datasets\footnote{\url{https://ogb.stanford.edu/docs/dataset_overview/}} to evaluate our framework and other baselines, since it comes with large, real-world graphs (million of nodes/edges) for realistic applications (i.e. network of academic, proteins). Moreover, it provides standard, open-sourced evaluation metrics and tools for benchmarking. 

Following the format of OGB, we design four prediction tasks of relations and higher-order patterns, and construct the corresponding datasets on heterogeneous graphs and hypergraphs\footnote{\url{https://www.cs.cornell.edu/~arb/data/}}. The original \texttt{ogb-mag} only contains features for `paper'-type nodes. We add the node embedding provided by \citep{yu2021heterogeneous} as raw features for the rest type of nodes in \texttt{MAG(P-A)/(P-P)}. For these four tasks, the model is evaluated by one positive query paired with certain number of randomly sampled negative queries, as listed in Table \ref{tab:mdata} along with other dataset statistics. The customized dataset for relation and higher-order pattern prediction is accessible via Box at \url{https://app.box.com/s/v9nszkai2gig13lm1o6q2ya82ap07gyb}.

\begin{table}[tp]
\centering
\caption{Dataset Statistics for Relation Prediction and Higher-order Pattern Prediction.}
\vspace{-2mm}
\label{tab:mdata}
\begin{center}
  \resizebox{0.48\textwidth}{!}{
  \begin{tabular}{lcccccccc}
    \toprule
    \textbf{Dataset}&\textbf{Query Type}&\textbf{\#Train (Pos.)}&\textbf{\#Val./Test (Pos.)}&\textbf{Pos./Neg.} \\
    \midrule
    \texttt{MAG(P-A)} & relation & 6,519,308 & 16,180 & 1:1000\\
    \texttt{MAG(P-P)} & relation & 5,199,201 & 22,639 & 1:1000\\
    \texttt{tag-math} & higher-order & 74,955 & 24,985 & 1:100\\
    \texttt{DBLP-coauthor} & higher-order & 79,566 & 26,522 & 1:1000\\
  \bottomrule 
\end{tabular}}
\end{center}
\vspace{-2mm}
\end{table}

\subsection{More Details about the Baselines} \label{apd:baseline}
For link prediction and relation prediction, we select baselines from the current OGB leaderboard \footnote{\url{https://ogb.stanford.edu/docs/leader_linkprop/}} based on two main factors: scalability and prediction performance. All the public models have code released with a technical report. With additional verification, we adopt the published numbers if available on the leaderboard. For the rest, we benchmark the model using their official implementation and tuning parameters as listed below.

\begin{itemize}
\item \textbf{Graph embedding}: graph embedding models for transductive learning such as Node2vec \cite{grover2016node2vec}, DeepWalk\footnote{\url{https://github.com/dmlc/dgl/tree/master/examples/pytorch/ogb/deepwalk}} \cite{perozzi2014deepwalk}. As implicit matrix factorization, it can be extensively optimized \cite{mohoney2021marius,xie2021demo} for large-scale graph mining tasks. The obtained node representation embeds the global position of target nodes in a given graph, which potentially can be exploited for link prediction.
\item \textbf{GCN family}: a graph auto-encoder model that using graph convolution layers to learning node representations, including GCN~\citep{kipf2016semi}, GraphSAGE~\citep{hamilton2017inductive}, and the derived models, such as Cluster-GCN~\citep{chiang2019cluster}, GraphSAINT~\citep{zeng2019graphsaint}.
\item \textbf{R-GCN}\footnote{\url{https://github.com/pyg-team/pytorch\_geometric/blob/master/examples}}~\citep{schlichtkrull2018modeling}: a relational GCN that models heterogeneous graphs with node/link types. 
\item \textbf{R-HGNN}\footnote{\url{https://github.com/yule-BUAA/R-HGNN}}~\citep{yu2021heterogeneous}: a heterogeneous GNN that focuses on learning relation-aware node representations with attention mechanism.
\item \textbf{SEAL}\footnote{\url{https://github.com/facebookresearch/SEAL_OGB}}~\citep{zhang2018link}: apply GCN with double radius labeling tricks to obtain subgraph-level readout for link prediction. SEAL reigns in the top spots of OGB leaderboard on multiple tasks, thanks to the expressiveness inherited from SGRL. The implementation we tested is specially optimized for OGB datasets provided in \cite{zhang2021labeling}.
\item \textbf{DE-GNN}\footnote{\url{https://github.com/snap-stanford/distance-encoding}}~\citep{li2020distance}: a provably more powerful SGRL that utilizes distance features (i.e. shortest path distance, landing probability) to assist GNNs in representing any set of nodes. DE-GNN can be applied to tasks such as node classification, link prediction and higher-order cases, with great performance.
\end{itemize}

For graph embedding approaches, we first use these models to generate node embeddings, and then train an MLP as the link predictor with input of the Hadamard product between hidden representations of two nodes. Then, as OGB guideline required, to perform data splitting, tune the MLP over the validation set, and  test it through the benchmark evaluator.

All canonical GNN baselines \footnote{\url{https://github.com/snap-stanford/ogb/tree/master/examples/linkproppred}} come with three message passing layers of 256 hidden dimensions, and a tuned dropout ratio in $\{0,0.5\}$ for full-batch training. Canonical GNN models combine node embeddings in the queried set as the representations of links/hyperedges, which are later fed into an MLP classifier for final prediction. Besides, they need to use full training data to generate robust node representations. The hypergraph datasets do not come with raw node features. Thus, canonical GNNs here use random features as input for training along with other model parameters. R-GCN and R-HGNN use relational \texttt{GCNConv} layers that support message passing with different relation types between nodes. The relation type of edges is used as the input beside node features.

SGRL-based models only use partial edges from the training set. Both SEAL and DE-GNN extract 1-hop enclosing subgraphs for training. SEAL applies three GCN layers of 32 hidden dimensions plus a sortpooling and several 1D convolution layers to generate readout of the target subgraphs for prediction. DE-GNN adopts shortest path distance calculated from each extracted subgraph as the input feature, and then employs two \texttt{TAGConv} layers of 100 hidden dimensions to generate readout of the queried node sets.

\subsection{Architecture and Hyperparameter \label{apd:model}}
\proj consists of a 2-layer MLP with ReLU activation for the embedding of node RPEs and a 2-layer RNN to encode joint walks obtained from queried subgraph joining. The hidden dimension of both networks is set to 64. Lastly, the concatenated hidden representations of queried node sets are fed into an 2-layer MLP classifier to make final predictions.

For link and relation prediction, we follow the inductive setting that only partial samples will be used for training. Over the training graph, we randomly select 5\% links as positive training queries, each paired with $k$-many negative samples ($k=50$ by default). We remove these links and use the remaining 95\% links to collect random walks and compute node RPEs via Algorithm \ref{alg:pdf}. For higher-order pattern prediction, we use the given graph before timestamp $t$ to obtain walks and RPEs, and then optimize model parameters by higher-order triplets provided in the training set. 

The results reported in Table \ref{tab:ogb} are obtained through the combination of hyperparameters listed in Table \ref{tab:hyper}. For the profiling of \proj in Table \ref{tab:cost_all} and Fig. \ref{fig:ttai}, we use the following combinations: \texttt{citation2} with $m=50,M=4,k=20$; \texttt{collab} with $m=2,M=200,k=20$, as relative small values of $m$, $M$, and $k$ already guarantee sufficient good performance. The rest hyperparameters remain the same as reported earlier.

\begin{table}[tp]
\caption{Hyperparameters Used for Benchmark \proj.}
\vspace{-2mm}
\label{tab:hyper}
\begin{center}
\resizebox{\columnwidth}{!}{
  \begin{tabular}{lccc}
    \toprule
    \textbf{Dataset} & \textbf{\#Steps $m$} & \textbf{\#Walks $M$} & \textbf{\#Neg. Samples $k$}  \\
    \midrule
    \texttt{citation2} & 4 & 200 & 50\\
    \texttt{collab} & 2 & 400 & 50\\
    \texttt{ppa}  & 4 & 200 & 50\\
    \texttt{MAG (P-A)} & 3 & 200 & 10\\
    \texttt{MAG (P-P)} & 4 & 100 & 50\\
    \texttt{tags-math}  & 3 & 100 & 10 \\
    \texttt{DBLP-coauthor} & 3 & 100 & 10  \\
  \bottomrule
\end{tabular}}
\end{center}
\vspace{-2mm}
\end{table}

\subsection{Computation Complexity Analysis}
We analyze the computation cost in the proposed framework \proj and identify three major parts:

\textbf{Random Walks}: the space complexity is $O(mM|\mathcal{V}|$), where $|\mathcal{V}|$ denotes the size of node set of the input graph; the time complexity is $O(mM|\mathcal{V}|/p)$, where $p$ is the number of threads. Practically, the number of steps $m$ generally lies in the range of $2 \sim 5$.

\textbf{Relative Positional Encoding}: the space complexity for RPE is $O(m^2M|\mathcal{V}|$) as there are at most $m\cdot M$ many distinct nodes in the set of walks originated from a single node. Meanwhile, the space requirement can be further reduced to 1/10 after pruning and remapping RPE via RPE-ID. RPE can be computed along with walk sampling. Thus, the time complexity of RPE computation is still $O(mM|\mathcal{V}|/p)$ by combining with random walks.

\textbf{Subgraph Joining}: for a query $Q$, the time complexity is $O(c\cdot mM|Q|/p)$ for joining all associated set of walks in a queried subgraph. $c$ is a scalar related to the size of $Q$. In practice, $|Q|=2$ for link prediction, $|Q| \ge 3$ for higher-order pattern prediction.

\subsection{Implementation Details}
We implemented our framework on top of PyTorch, NumPy, and OpenMP. \texttt{uhash} is adopted to serve light and high efficient indexing for RPEs associated with sampled walks. For better parallelization, the computationally intensive part is written in C language with bindings to support Python APIs, including walk sampler, RPE encoder, and subgraph joining operation. To reduce the overhead of hybrid programming, we use RPE-IDs and native C/Numpy arrays instead of Python objects to exchange results between API calls and underlying C functions. We also provide Python APIs to support customizing above-mentioned parallel operations. The \proj framework is open-source at \url{https://github.com/Graph-COM/SUREL} and free for academic use under the BSD-2-Clause license.

\end{document}